%% file: main.tex
\documentclass[10pt,journal,compsoc]{IEEEtran}
\usepackage{times}
\usepackage{graphicx}
\usepackage{amsmath}
\usepackage{booktabs}
\usepackage{multirow}
\usepackage{xcolor}
\usepackage{amssymb}
\usepackage{diagbox}
\usepackage{subfigure}
\usepackage{graphicx}

\usepackage{caption}
\usepackage{marvosym}
\usepackage[pagebackref,breaklinks,colorlinks]{hyperref}

\ifCLASSOPTIONcompsoc
  \usepackage[nocompress]{cite}
\else
  \usepackage{cite}
\fi
\ifCLASSINFOpdf
\else
\fi
\begin{document}
%
\title{\textit{SketchMetaFace}: A Learning-based Sketching Interface for High-fidelity 3D Character Face Modeling}
%
%
%
%

\author{Zhongjin Luo,
        Dong Du,
        Heming Zhu,
        \\ Yizhou Yu,~\IEEEmembership{Fellow,~IEEE,}
        Hongbo Fu,
        Xiaoguang Han\textsuperscript{\Letter},~\IEEEmembership{Member,~IEEE}
\IEEEcompsocitemizethanks{\IEEEcompsocthanksitem Z. Luo, D. Du, H Zhu, and X. Han are with the School of Science and Engineering, The Chinese University of Hong Kong, Shenzhen, China. Y. Yu is with the Department of Computer Science at the University of Hong Kong. H. Fu is with the School of Creative Media, City University of Hong Kong.
\IEEEcompsocthanksitem X. Han is the corresponding author. E-mail: hanxiaoguang@cuhk.edu.cn}
\thanks{Manuscript received xxxx.}}

%
%

\markboth{Journal of \LaTeX\ Class Files,~Vol.~xx, No.~x, xxxx}%
{Shell \MakeLowercase{\textit{et al.}}: Bare Demo of IEEEtran.cls for Computer Society Journals}
%



\IEEEtitleabstractindextext{
\begin{abstract}
Modeling 3D avatars benefits various application scenarios such as AR/VR, gaming, and filming. Character faces contribute significant diversity and vividity as a vital component of avatars. However, building 3D character face models usually requires a heavy workload with commercial tools, even for experienced artists. Various existing sketch-based tools fail to support amateurs in modeling diverse facial shapes and rich geometric details. In this paper, we present \textit{SketchMetaFace} - a sketching system targeting amateur users to model high-fidelity 3D faces in minutes. We carefully design both the user interface and the underlying algorithm. First, curvature-aware strokes are adopted to better support the controllability of carving facial details. Second, considering the key problem of mapping a 2D sketch map to a 3D model, we develop a novel learning-based method termed ``Implicit and Depth Guided Mesh Modeling'' (IDGMM). It fuses the advantages of mesh, implicit, and depth representations to achieve high-quality results with high efficiency. In addition, to further support usability, we present a coarse-to-fine 2D sketching interface design and a data-driven stroke suggestion tool. User studies demonstrate the superiority of our system over existing modeling tools in terms of the ease to use and visual quality of results. Experimental analyses also show that IDGMM reaches a better trade-off between accuracy and efficiency. \textit{SketchMetaFace} is available at \url{https://zhongjinluo.github.io/SketchMetaFace/}.
\end{abstract}
}

\maketitle

\IEEEdisplaynontitleabstractindextext

%
\IEEEpeerreviewmaketitle

\input{section/introduction}
\input{section/related_works}

\input{section/ui}
\input{section/algorithm}
\input{section/results}
\input{section/conclusion}
\ifCLASSOPTIONcaptionsoff
  \newpage
\fi



%



\bibliographystyle{IEEEtran}
\bibliography{reference}

\vfill







\end{document}

%% file: section/introduction.tex
\IEEEraisesectionheading{\section{Introduction}\label{sec:introduction}}

\input{figure/fig_teaser}

\IEEEPARstart{C}{reating} 3D virtual avatars is a prolonged research topic in computer graphics and benefits various usage scenarios such as filming, gaming, and art designing. Typically, this process is a highly skilled task, as experienced artists need to spend several days or even months formally sculpting high-fidelity 3D faces with vivid surface details using commercialized 3D modeling tools (e.g., ZBrush, MAYA, and 3D MAX). To assist amateur users in freely instantiating their ideas as professional modelers, researchers in computer graphics and human-computer interaction have designed systems that allow users to model 3D shapes with freehand sketches based on geometric principles~\cite{igarashi1999teddy,nealen2007fibermesh,Borosan:2012:RAR,sykora2014ink,pan2015flow,li2017bendsketch}. Although traditional sketch-based 3D modeling systems, such as Teddy~\cite{igarashi1999teddy} and FiberMesh~\cite{nealen2007fibermesh}, enable amateur users to create 3D models, they usually require tremendous manual work to specify complex geometry. 

Thanks to the recent progress in deep learning, the understanding of freehand sketches and the quality of single-view generation have reached an unprecedented level. Several intelligent sketch-based modeling systems have been developed to enable novice users to create visually plausible 3D models within a few minutes~\cite{zhong2020deep,zhong2022study,xu2022deep}. Closest to our work, DeepSketch2Face~\cite{han2017deepsketch2face} presents the first deep learning-based sketching system for modeling 3D faces by mapping 2D sketches into a parametric space for face generation. However, considering the limited representation power of the parametric model, DeepSketch2Face can only produce 3D human faces with fixed styles and cannot be used for sculpting expressive skin wrinkles. SimpModeling~\cite{luo2021simpmodeling} proposed a two-phase scheme that allows for diverse animalmorphic head modeling using 3D sketches. Nevertheless, it is challenging for users to work with as it relies on complicated and user-unfriendly 3D interactions. Additionally, the system struggles to generate fine-grained details due to the ambiguity of mono-typed strokes and the limited capability of PIFu~\cite{saito2019pifu,saito2020pifuhd}.

In this paper, we design and present \textit{SketchMetaFace}, a powerful sketch-based 3D face modeling system that addresses the following challenges:

\textit{\textbf{Accuracy}.} Recent learning-based sketching systems~\cite{han2017deepsketch2face,li2018robust,li2020sketch2cad,du2020sanihead,luo2021simpmodeling} allow novice users to create visually-plausible 3D models with a few strokes. However, they are not capable of designing shapes with fine-grained details. To assist users in conveying their ideas more accurately, we adopt curvature lines~\cite{iarussi2015bendfields,li2017bendsketch,li2018robust} in learning-based 3D face modeling. 
We will demonstrate how the curvature-aware strokes significantly boost the quality of detailed surfaces generated from sketches.

Although existing models~\cite{lun20173d,delanoy20183d,wang20233d,wang2018pixel2mesh,Guillard_2021_ICCV} can map 2D images, including sketch images, to 3D shapes, they may fail to generate watertight 3D meshes with delicate details. A straightforward way to produce shapes with surface details is to blend high-quality multi-view depth maps generated by image translation~\cite{isola2017image}. Nonetheless, it is nontrivial to fuse the generated depth maps seamlessly into a watertight mesh. An alternative approach is to adopt the pixel-aligned implicit function (PIFu)~\cite{saito2019pifu,saito2020pifuhd} to reconstruct watertight 3D shapes from single images. However, PIFu exhibits bounded performance in generating high-fidelity geometric details. Inspired by the fact that the depth map produced by image translation contains more intriguing details than PIFu-generated shapes, we propose IDGMM, i.e., \textbf{I}mplicit and \textbf{D}epth \textbf{G}uided \textbf{M}esh \textbf{M}odeling. It enjoys the merits of mesh, depth-map and implicit representations to produce high-quality 3D shapes from curvature-aware sketch images.

\textit{\textbf{Usability}.} While curvature-aware strokes empowers users to create 3D faces with fine-grained details, it may increase their cognitive load. To address this issue, we interview potential users and thoroughly analyse their requirements. We design our system based on the analyzed requirements and formulate a coarse-to-fine interactive scheme: users first get used to the system with mono-typed sketches and then switch to fine-detail crafting with curvature-aware strokes soon as users get familiar with the system. We also carefully design a stroke suggestion component that bridges the gap between coarse and detailed sketching. Moreover, to follow the ``as-2D-as-possible" principle, we keep the placement of ears as the only 3D interaction in our system.

To demonstrate the effectiveness of our system, we carefully conduct user studies, from which we conclude that our proposed system exhibits better usability than existing sketch-based 3D face modeling systems~\cite{han2017deepsketch2face,luo2021simpmodeling}. Our system allows amateur users to create diverse shapes with fine-grained geometric details. By conducting comparisons against existing inference algorithms for mapping a single sketch to a 3D model, we demonstrate that results generated by our proposed IDGMM better reflect the appearances of the input sketches. Ablation studies are conducted to justify each design in our interface and algorithm. The contributions of our paper can be summarized as follows:
\begin{itemize}

\item We present a novel, easy-to-use sketching system that allows amateur users to model high-fidelity 3D character faces in minutes (as seen in Fig.~\ref{fig:teaser}).

\item We carefully design a 
user interface: 1) the face modeling work follows a coarse-to-fine scheme and relies mainly on intuitive 2D freehand sketches; 2) we adopt curvature-aware strokes for modeling geometric details; 3) we introduce a data-driven suggestion tool to ease the cognitive load throughout the sketching process.

\item We propose a novel method, i.e., Implicit and Depth Guided Mesh Modeling (IDGMM), which fuses the advantages of mesh, implicit, and depth representations for detailed geometry inference from 2D sketches.

\end{itemize} 

%% file: figure/fig_teaser.tex
\begin{figure*}[htbp]
\centering
  \includegraphics[width=.93\textwidth]{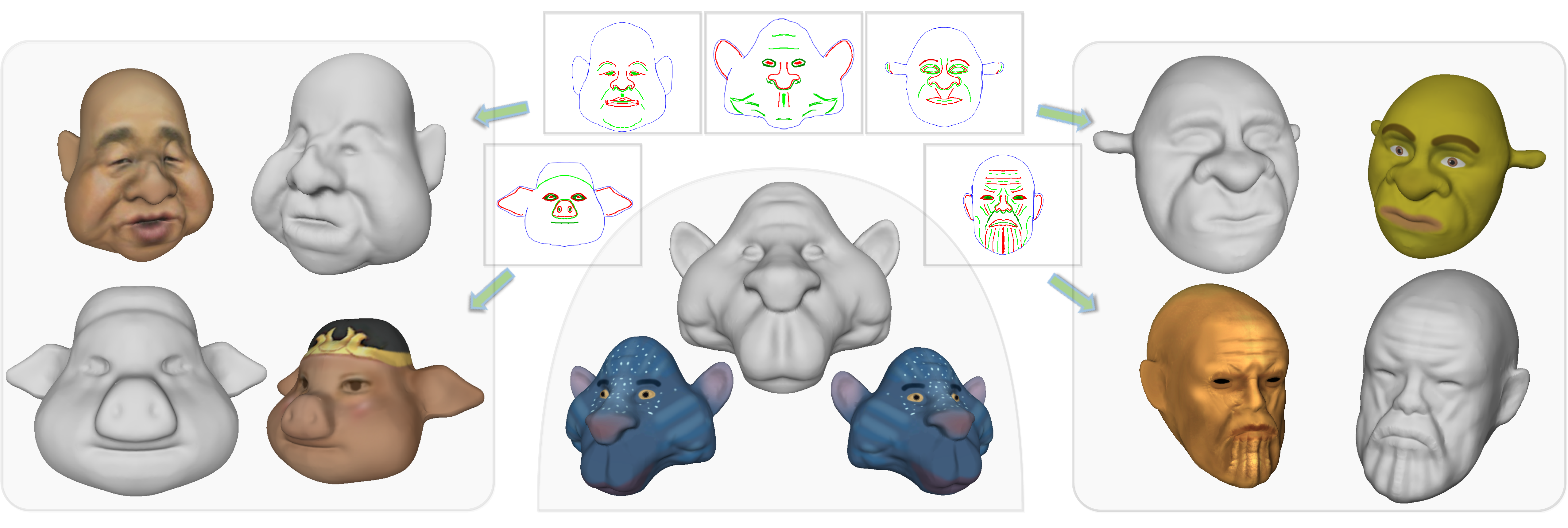}
  \caption{We present \textit{SketchMetaFace}, a novel sketching system designed for amateur users to create high-fidelity 3D character faces. With curvature-aware strokes (valley strokes in green and ridge strokes in red), novice users can smoothly customize detailed 3D heads. Note that our system only outputs geometry without texture and texturing is achieved using commercial modeling tools.}
  \label{fig:teaser}
\end{figure*}

%% file: section/related_works.tex
\section{Related Work}
In this section, we will present relevant studies on 3D avatar modeling, geometrical sketch-based modeling, and data-driven sketch-based modeling. We are aware of the breathtaking progress in sketch-based 2D image generation of faces~\cite{deepfacedeiting,deepfacedrawing}. However, we will not discuss these in detail due to the page limit. 

\subsection{3D Face from 2D Image} 
Creating visually plausible 3D avatars is a long-standing computer graphics and vision problem. Compared with 3D face reconstruction methods, which take multi-view~\cite{xiao2022detailed, Bai_2020_CVPR} or monocular video~\cite{mbr_CFML,ctface} as input, single image reconstruction (SVR) and sketch-based modeling provide more casual means for novices to customize 3D faces. Single-image 3D face reconstruction can be roughly divided into two streams, namely, photo-realistic human face reconstruction and caricature face reconstruction.

The works on single-image photo-realistic face reconstruction can be further separated into two genres, i.e., parametric and shape-from-shading methods. However, neither can be directly adopted for modeling detailed 3D faces. Parametric-based models~\cite{cao2013facewarehouse,deng2019accurate,tran2018nonlinear} fall short in representing shapes with novel and customized surface details. Shape-from-shading-based methods~\cite{richardson2017learning,t2018extreme} suffer from deriving geometric clues from non-photo-realistic image inputs, e.g., caricature images and sketches.

Compared with single-image realistic 3D faces generation, which has been extensively studied and achieved exceptionally high quality, the researches on 3D caricature avatars are relatively sparse. A possible reason is that caricature 3D faces are shapes with more diversified geometry, making them extremely hard to be regularized into a single parametric model losslessly. Some work \cite{liu2009semi,zhang2020landmark,wu2018alive} introduced deformations to increase the capability of parametric models. However, their works are still far from generating high-fidelity 3D caricature shapes of various styles. More importantly, given that most single-image caricature face modeling methods require high-quality images as input, novice users cannot further customize the shape as they wish.

Recently, researchers have also explored various schemes for interactive modeling from 2D sketch images~\cite{delanoy20183d, lun20173d, wang20233d, zhang2021sketch2model, chowdhury2022garment, han2017deepsketch2face, du2020sanihead}. In line with our work, DeepSketch2Face~\cite{han2017deepsketch2face} proposed a sketch modeling system that allows users to create caricature heads from scratch. Their method relies on a CNN-based model to parse a user-drawn sketch as the parameters for a morphable face model. However, since the 3D caricature shape is confined to the parametric caricature face model, DeepSketch2Face cannot faithfully reflect  large deformations and wrinkle details presented in the sketch. To address this issue, SAniHead~\cite{du2020sanihead} proposed a view-surface collaborative mesh generative network, which turns dual-view freehand sketches into animalmorphic heads. Nevertheless, it fails to synthesize novel shapes deviating from training datasets due to the restricted generalization ability of their network. Our system utilizes the advantages of mesh, depth-map, and implicit representations to generate high-quality 3D shapes from curvature-aware sketch images.

\subsection{Geometrical Sketch-based Modeling}
Designing free-form 3D shapes via freehand sketching has drawn considerable attention in recent decades~\cite{ding2016survey}. Igarashi et al.~\cite{igarashi1999teddy} pioneer by proposing the first sketch-based modeling system that allows users to create 3D shapes from scratch by sketching 2D contour lines. A large stream of subsequent researches~\cite{karpenko2006smoothsketch, schmidt2007shapeshop, bernhardt2008matisse, joshi2008repousse, gingold2009structured, olsen2011naturasketch} has mainly focused on designing novel interpolation functions to interpolate sketched contours lines smoothly. Unlike the sketch-based modeling systems mentioned above, which take 2D sketches as input, Fibermesh~\cite{nealen2007fibermesh} allows users to model free-form surfaces by sketching and manipulating 3D curves. While Fibermesh~\cite{nealen2007fibermesh} and its follow-up systems~\cite{bae2008ilovesketch, schmidt2009analytic} reduce the ambiguity remarkably with explicitly defined 3D curves, they are not capable of or are not friendly for novice users to carve organic surface details (e.g., skin wrinkles). 

To emboss interpolated surfaces with sharper details, various methods introduce sketches with different semantics~\cite{shao2012crossshade, xu2014true2form} or curvature cues~\cite{iarussi2015bendfields, li2017bendsketch} to formulate more determined constraints. However, additional inputs may significantly increase novice users' cognitive load. Inspired by BendSketch~\cite{li2017bendsketch}, our system allows users to draw with curvature-aware strokes, which serve as a less ambiguous means for users to specify the bulge and sink on faces accurately. To reduce the cognitive load of using curvature-aware strokes, we introduce a carefully designed sketch suggestion module to support amateurs in getting familiar with our system intuitively.

\subsection{Data-driven Sketch-based Modeling}
The recent decade has witnessed the emergence of data-driven methods for sketch-based 3D shape generation thanks to large-scale 3D datasets. The data-driven sketch-based modeling systems can be roughly divided into two streams regarding the shape generation approaches, i.e., retrieval-based and learning-based.

Retrieval-based methods~\cite{eitz2012sketch,li2016retrival,qi2021toward,luo2022structure} consume a freehand sketch for the query and search for the most similar shape from the data warehouse as the reconstruction output. Fan et al.~\cite{fan2013modeling} propose a suggestive interface with shadow guidance to guide object sketching. However, shadow guidance may introduce severe visual cluttering for sketches with different semantics. Xie et al.~\cite{xie2013sketch} proposed to retrieve candidate object parts from a database with part sketches for further assembly. Recently, deep neural networks have been applied for retrieval-based sketch modeling systems~\cite{wang2015sketch}, which have shown their superiority compared to their traditional learning-based counterparts in handling noisy sketch input created by novice users. However, limited by the capacity of the data warehouse, retrieval-based sketch modeling may produce shapes that drift away from input sketches.


In recent years, learning-based solutions have been popular for sketch-based 3D shape generation and editing~\cite{lun20173d,delanoy20183d,wang20233d,zhong2020towards,zhang2021sketch2model,cheng2022cross,kong2022diffusion,su2018interactive,huang2016shape,han2017deepsketch2face,luo2021simpmodeling,du2020learning,li2020sketch2cad,chowdhury2022garment,nishida2016interactive}. For example, Nishida et al.~\cite{nishida2016interactive} proposed inferring urban building parameters from freehand sketches with convolutional neural networks, while Huang et al.~\cite{huang2016shape} presented an interactive modeling system that infers parameters for procedural modeling from sketches. DeepSketch2Face~\cite{han2017deepsketch2face} proposed a deep regression model that converts a sketch into the parameters of a morphable 3D caricature face model. However, the above parametric regression-based methods work only for 3D shapes within a specific category that can be easily parameterized. Du et al.~\cite{du2020learning} adopted implicit learning to produce artificial object parts from sketches and proposed a deep regression model to predict the position of the parts, while Sketch2CAD~\cite{li2020sketch2cad} enables users to achieve controllable part-based CAD object modeling by sketching in context. SimpModeling~\cite{luo2021simpmodeling} utilized a coarse-to-fine modeling scheme, allowing users to create desired animalmorphic heads with 3D curves and on-surface sketching. We argue that 2D sketching would be more intuitive than 3D sketching since most novice users are more familiar with 2D interfaces and interactions. Furthermore, SimpModeling falls short in generating fine-grained geometric details due to the ambiguity of mono-typed strokes and the bounded capability of its shape-generation network. In this paper, our system allows users to create 3D high-fidelity facial models with 2D curvature-aware sketches intuitively.


%% file: section/ui.tex
\section{User Interface}
\label{ui}
This section first summarizes the requirements of designing sketch-based modeling for novice users to customize high-fidelity faces of highly diversified styles. On top of the design goals, we will introduce the crucial designs of our system and justify how they reflect the design goals. Please refer to the accompanying video for sketch-based modeling in action.

\input{figure/fig_ui_pipeline}

\subsection{Design Requirements and Analysis}
\label{requirements}
In the design process of our sketch-based 3D face modeling system, we interviewed 11 participants with different levels of modeling experience to analyze the demands for a user-friendly sketch-based modeling interface. Three of these participants were modelers with more than five years of experience in 3D modeling, while the rest were novice users with no or little knowledge of 3D modeling. Based on the responses, we summarize the following design goals and the corresponding design choices for our system:

\noindent{\textit{\textbf{Coarse to Fine (R1).}}} After briefly introducing the background knowledge about sketch-based 3D shape modeling, we first discuss whether users prefer working in a top-down or bottom-up manner. All experts and most novice users preferred to model the shape in a top-down manner. Therefore, our proposed sketch-based modeling system allows users to model 3D faces in a coarse-to-fine manner~\cite{luo2021simpmodeling}. In the coarse stage, users can design the contour and the attachment of the faces (e.g., ears). After users finish designing a coarse head shape, they will move on to the fine-grained shape modeling stage, where they can carve geometrical details such as wrinkles, mouths, eyes, etc. Note that we treat ears as attachments and adjust their position through 3D interactive operations in the coarse stage since it is difficult to determine the 3D location of attachments just via frontal-view sketching.

\noindent{\textit{\textbf{As 2D as Possible (R2).}}} When discussing whether 3D interactions should be dominant in the system, most novice users mentioned that they prefer to express their ideas through 2D drawings. Interestingly, even professional modelers agree that 2D interactions should be the dominant interaction for the system, as they believe novices may get bored with manipulating the cameras and the 3D shapes. To this end, our system follows the ``as-2D-as-possible'' principle. Users can finish most of the design only with a 2D sketch pad, and 3D interactions (e.g., tuning the layout of ears) are introduced only when necessary.

\noindent{\textit{\textbf{Agile and Precise (R3).}}} While some amateurs mentioned that they want to carve a 3D face carefully according to a reference character face, others only intend to customize a visually-plausible 3D face with a few strokes. Hence, our system allows users to customize 3D faces with different degrees of interaction complexity, as shown in the demo video. Novice users can quickly orchestrate a visually plausible 3D face with the dedicated sketch stroke suggestion module. The sketch stroke suggestions also serve as a decent initialization for detailed face modeling. For users who are interested in carving customized surface details, we provide curvature-aware strokes that allow the specification of surface details to be more precise. 

\subsection{Coarse Shape Modeling}
\label{coarseshape}
To support the design requirements mentioned in Section \ref{requirements}, in our system, the modeling of high-fidelity 3D faces is decomposed into coarse shape modeling and fine detail sketching (\textbf{R1}). Users may start designing a coarse 3D face by drawing face contour lines on the 2D sketching pad view, as illustrated in Fig.~\ref{fig:ui_pipeline}. Novice users could switch to the symmetric sketching mode. Under this mode, mirror-symmetrical strokes will be generated as the user draws on the sketch pad. In this stage, our system can produce a 3D model in a real-time manner by responding to each drawing operation.

\noindent{\textit{\textbf{Profile Depth Editing.}}} The essence of our system lies in eliminating 3D user interactions (\textbf{R2}). However, the generated 3D faces with single-view contour strokes lack depth variances along the z-axis due to the missing constraints on the depth channel. To this end, we deliberately design a profile depth editing interaction scheme that allows users to specify the face contours in the lateral view. Once users switch to the depth editing mode, a new canvas will appear with an initial side-view rendered 3D face contour. As seen in Fig.~\ref{fig:ui_pipeline}, novice users may design shapes with sharp-variant depth by revising the profile sketch without directly manipulating the 3D shapes.

\noindent{\textit{\textbf{Ear Modeling.}}} The attachments of 3D faces, i.e., the ears, play an essential role in shaping a virtual character's characteristics and styles. Unlike nose, eyes, and mouth, ears (and other face attachments) are of greater diversity in 3D layout, making it challenging to use only frontal-view sketching to express. To this end, our system uses separate meshes to represent the face and the ears for better expressiveness. Users may customize the ears by drawing their contour lines on the 2D sketch pad view, like specifying the coarse head shape. Specifically, the ears (also for other attachments like horns) are sketched on individual canvas layers, which facilitate users to manipulate their 2D attachment layouts and help the backend models learn diversified attachment shapes. As illustrated in Fig.~\ref{fig:ui_pipeline}, users can modify the 3D layout of the ears in the 3D view for more precise control of the generated shape. Users can also copy attachments as in RigMesh~\cite{Borosan:2012:RAR}. It is worth mentioning that layout manipulation and attachment copying are the only 3D operations in the whole modeling procedure (\textbf{R2}).


\subsection{Fine Detail Sketching}
\label{finedetails}
After the user customizes the coarse face shape, they may further characterize the detailed facial geometry, e.g., eyes, noses, mouth, and wrinkles. Although previous works, e.g., DeepSketch2Face~\cite{han2017deepsketch2face} and SimpModeling~\cite{luo2021simpmodeling}, allow users to edit surface details through 2D and 3D sketching, they fall short in generating diversified and controllable surface details due to the ambiguous mono-typed sketch strokes. 

\noindent{\textit{\textbf{Curvature-aware Strokes.}}} We adopt curvature-aware strokes~\cite{li2017bendsketch} to alleviate the sketch's ambiguity, enabling users to carve surface details precisely (R3). Specifically, two types of strokes (i.e., ridge and valley) are defined. Before each stroke drawing, the user needs to pick a type first. Different stroke types are visualized with different colors (i.e., red for ridge and green for valley). Our system also supports tunable depth for each stroke, which defines the curvature amplitude, i.e., greater depth (darker color) means a higher ridge or deeper valley. 


\input{figure/fig_suggestion.tex}

\noindent{\textit{\textbf{Stroke Suggestions.}}} While curvature-aware strokes significantly improve the controllability of our system, they inevitably bring additional cognitive load for novice users. To address this, we carefully design a data-driven stroke suggestion tool. Consider a scenario when a user wishes to draw a pig nose on the face, as illustrated in Fig.~\ref{fig:suggestion}. Our system allows the user to pick the ``nose'' type and select a ``pig'' style first, and then draw a contour to specify the rough shape and the location where they wish to place the nose. After that, a set of strokes with the specified category, as well as the corresponding shapes, is retrieved from the database and presented as ``Suggestion''. The user can picks one which can be placed automatically or after manually adjusting the location and size. Users were provided 20 suggestions each time, and the retrieved sketches are editable. With such a suggestion tool, amateurs can quickly compile a neat 3D face model with the high-quality sketch strokes in the database and kick off instantiating their ideas on a decent basis. The suggestion tool is implemented by a retrieval neural network based on the auto-encoder structure, please refer to the supplemental materials for details.

\noindent{\textit{\textbf{Instant Shape Preview.}}} An instant preview of the 3D shape could serve as guidance for further sketching. However, due to the geometry complexity, the model inference in the stage of fine-detail sketching takes around 0.5s, making it unable to support real-time response. Our video shows that we adopt image space rendering and generate the frontal-view normal map as a real-time shape preview. Please refer to the supplemental materials for the implementation details of the instant preview module.

%% file: figure/fig_ui_pipeline.tex
\begin{figure*}[htbp]
  \centering
  \includegraphics[width=.95\linewidth]{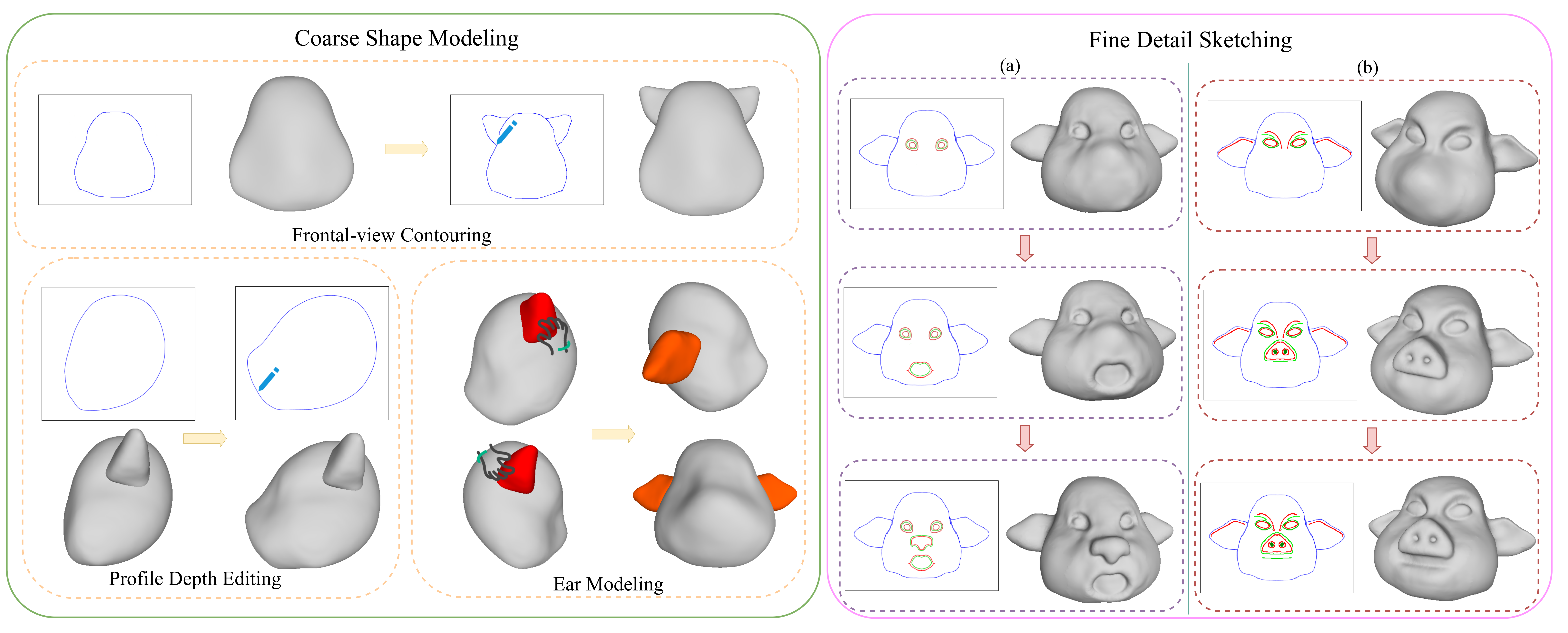}
  \caption{An illustration of the interactions supported by our system. In the \emph{Coarse Shape Modeling} stage, users may define coarse 3D faces with frontal-view contouring, profile depth editing, and ear modeling. In the \emph{Fine Detail Sketching} stage, users can further carve fine-grained surface details with the curvature-aware strokes.}
  \label{fig:ui_pipeline}
\end{figure*}

%% file: figure/fig_suggestion.tex

\begin{figure}[htbp]
  \centering
  \includegraphics[width=.98 \linewidth]{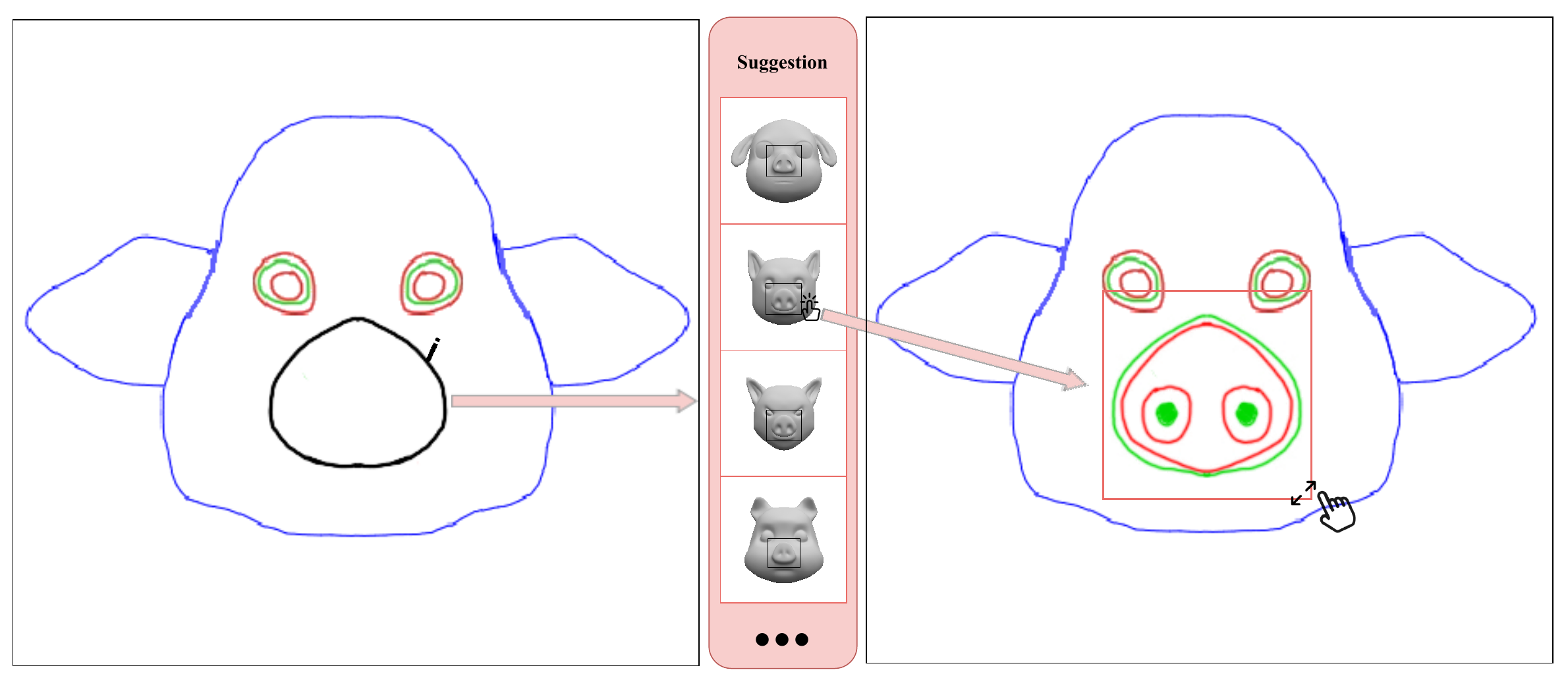}
  \caption{An illustration of our stroke suggestion component. Soon after users specify the style, target region, and facial components to be modeled, the stroke suggestion component retrieves the relevant curvature-aware strokes. Users may also manipulate the layout for the retrieved strokes through dragging and scaling.}
  \label{fig:suggestion}
\end{figure}

%% file: section/algorithm.tex
\section{Methodology}
\label{sec:algorithm}

In this section, we present the details of the backend models that support the interactive sketching interface.

\noindent{\textit{\textbf{Overview.}}} Following our coarse-to-fine interface design, we discuss the algorithms used for the two stages accordingly. In the coarse stage, as illustrated in Fig.~\ref{fig:pipeline_coarse}, we propose a part-separated implicit learning method that maps the coarse input sketch $S_r$ to separated part meshes (i.e., face and attachments). After the user tunes the part layout, these separated meshes are merged into a single mesh $M_c$. We then render the outer contour~\cite{decarlo2003suggestive} of $M_c$ into the sketch image $S_c$, on which users can add fine strokes in the detail sketching stage.

\input{figure/fig_pipeline_coarse.tex}

In the detail sketching stage, users may further craft fine-grained surface details through sketching on the rendered coarse sketch image $S_c$. To generate detailed geometry $M_f$ from the fine sketch $S_f$,  as shown in Fig.~\ref{fig:pipeline_fine}, we propose IDGMM, which learns a progressive deformation from $M_c$ to $M_f$, under the guidance of both the learned implicit field (SDF) and the learned depth map from $S_f$.

\input{figure/fig_pipeline_fine}

\subsection{Preliminary}
Before introducing the proposed {model}, we will briefly review some relevant concepts and building blocks.

\noindent{\textit{\textbf{Pix2Pix.}}} Given a source image $I_s$, Pix2Pix~\cite{isola2017image} learns a 
mapping from $I_s$ to a target image $I_t$, i.e., $f: I_s \to I_t$ in an adversarial manner. Commonly, a U-Net is adopted to model this translation, and the conditional GAN loss and the reconstruction loss ($L_1$ or $L_2$ loss) are used for training. In our model, the Pix2Pix module is adopted for translations among sketch images, depth maps, and normal maps.


\noindent{\textit{\textbf{Implicit Learning.}}} Recently, various deep representations have been used for 3D shape reconstruction, e.g., voxels, point clouds, meshes, and implicit fields. Among them, implicit field-based methods achieve state-of-the-art performance~\cite{mescheder2019occupancy, park2019deepsdf, chen2019learning}. There are two commonly used formulations to model implicit surfaces: occupancy and signed distance function (SDF). Occupancy is a continuous function $g_o$ that maps a query point $p \in R^3$ to a binary status $o \in \{0, 1\}$, indicating inside/outside of the surface. SDF is a function $g_s$  that maps $p$ to its signed distance $s$ to the underlying surface. A multi-layer perception (MLP) is usually adopted for approximating $g_o$ or $g_s$.

\noindent{\textit{\textbf{PIFu.}}} Among the works relevant to single image 3D shape reconstruction, pixel-aligned implicit function (PIFu) outperforms its counterparts in generating results better matching input images. Specifically, PIFu models a function $h$ to map $p \in R^3$ with a projected pixel-aligned feature $f_p$ to an occupancy $o$ or SDF value $d$, i.e., $h: \left\{p, f_p\right\} \to o/d$. Firstly, an hourglass architecture~\cite{newell2016stacked} is applied on $I$ to obtain a feature map $I_f$. Then, $p$ is projectd onto $I_f$ to obtain $f_p$. MLP is used to model $h$. Our system also requires input-aligned results, so we adopt PIFu as the base module for shape inference from sketch images. Our method uses SDF-based PIFu since it is more suitable for providing deformation guidance. 
 
\noindent{\textit{\textbf{PIFu with Normal Input.}}} As a follow-up work of PIFu, PIFuHD~\cite{saito2020pifuhd} proposed a coarse-to-fine pixel-aligned implicit shape learning pipeline to generate more geometry details. More specifically, it utilizes PIFu as the coarse-level learning and adopts generated normal maps for fine-level learning. Inspired by PIFuHD, we infer normal maps from the input sketch images with Pix2Pix to assist in learning fine-grained surface details. Similar to the design proposed in PIFuHD, we maintain a tiny MLP that extracts local image features from the inferred normal maps to generate high-frequency details. In the following sections, we will use PIFu-N to denote our PIFu with normal input.



\subsection{Coarse Modeling}
In the coarse stage, users only need to draw a rough outline for a desired face, i.e., the face contour and attachment contours (e.g., ears). A straightforward way to generate a coarse model from the outline sketch $S_r$ is to use PIFu, which maps $S_r$ to an implicit field.  Subsequently, Marching Cubes~\cite{lorensen1987marching} can be adopted to extract a mesh from the implicit field. However, as the attachments and the face are represented with a single mesh, users cannot directly manipulate the layout for the attachments, thus significantly weakening users' control over modeling results.

\subsubsection{Part-separated PIFu} To boost the controllability of our system, we present a novel part-separated PIFu. Let's first consider a common scenario where a face contains a left ear and a right ear. As shown in Fig.~\ref{fig:pipeline_coarse}, three different PIFu modules are used to model the three parts separately. They use different MLPs but share a common encoder that maps $S_r$ to feature maps. In our implementation, the number of parts is fixed. The MLPs designed for ear parts can also be used to generate ear-like attachments, such as horns. 

The 3D location of each ear is kept without any normalization during training, which makes the network learn the layout of ears automatically. After obtaining the implicit field of each part, we extract separated meshes from them (for better efficiency, $64^3$ resolution is used for marching cube). After users manipulate 3D ear placements, those meshes are merged into a single one with a \emph{corefine-and-compute-union} operation provided by CGAL~\footnote{CGAL: the Computational Geometry Algorithms Library. https://www.cgal.org/.}. After this step, we apply a remeshing method~\cite{botsch2004remeshing} to get $M_c$.

Although our curvature-aware strokes contain a ``depth'' attribute for depth controlling, it can only model local depth. Thus we provide a profile sketching tool for global depth editing (as seen in Fig.~\ref{fig:ui_pipeline}). Specifically, the profile contour is treated as the handle to define a Laplacian deformation~\cite{sorkine2004laplacian}. Since $M_c$ in the coarse stage is in a low resolution, the Laplacian deformation can be performed in real-time.

\subsubsection{Training}
The part-separated PIFu is trained in a fully-supervised manner. For each character face mesh $M$ in the dataset, we render its contours as a sketch image input. To prepare the ground truth data for training our part-separated PIFu used in the coarse stage, we smooth faces meshes $M$, and then segment them into distinct parts (i.e., faces and attachments). The ground-truth SDF values for each part are calculated in the world coordinates. During training, we use the $L_1$ metric to measure the difference between the predicted SDF values and the ground truth.


\subsection{IDGMM: Implicit and Depth Guided Mesh Modeling}
In the fine stage, $M_c$ is first rendered into a new contour map $S_c$. Then users will draw curvature-aware strokes over $S_c$, and we denote the updated sketch image as $S_f$. This section discusses the method to map $S_f$ to a model denoted as $M_f$. It resembles the shape of $S_c$ but contains local geometric details reflected by $S_f$, as illustrated in Fig.~\ref{fig:pipeline_fine}. 

Recently, many deep-learning-based methods~\cite{mescheder2019occupancy, park2019deepsdf, chen2019learning, saito2019pifu, saito2020pifuhd} have been proposed to map a sketch image to a 3D model. A straightforward solution is to apply PIFu-based methods~\cite{saito2019pifu, saito2020pifuhd} and extract the surface mesh with Marching Cubes (MC)~\cite{lorensen1987marching}. However, MC is time-consuming (5 seconds $256^3$ iso-value grid) when extracting high-resolution surfaces and fails to meet the requirements for interactive editing. To this end, we apply the field-guided deformation formula to speed up the extraction of detailed surfaces from implicit fields.

Specifically, our method takes $M_c$ and $S_f$ as input and learns the displacement for each vertex on $M_c$ with the help of both the implicit and depth-map representations. Before conducting deformation, we subdivide the regions~\cite{botsch2004remeshing} where detail strokes are drawn to better afford geometric details. Note that the sketching process is iterative, and the input $M_c$ at the n-th step is the resulting mesh at step (n-1). For simplicity, we still use $M_c$ to represent the input coarse mesh of each step.

\subsubsection{Implicit-guided Mesh Updating} 
Inspired by the work~\cite{sharf2006competing}, SimpModeling~\cite{luo2021simpmodeling} proposed a strategy for mesh deformation under the guidance of implicit fields, but it is inefficient: 1) SimpModeling utilizes an occupancy field and needs to determine the updated vertices by a dense sampling way; 2) to stabilize the deformation, the Laplacian deformation technique~\cite{sorkine2004laplacian} is adopted. 

In contrast, we update $M_c$ directly with the guidance of the continuous SDF field to keep robustness during deformation, which dramatically reduces the computational cost of the Laplacian deformation (i.e., updating each vertex $\mathbf{v} \in M_c$ via $\mathbf{v'} = \mathbf{v}+g_{s}(\mathbf{v})\mathbf{n}$, where $\mathbf{n}$ indicates the normal of $\mathbf{v}$ and $g_{s}(\mathbf{v})$ is the SDF value of $\mathbf{v}$). The above updating mechanism could be performed iteratively for multiple times, but its enhancement was slight. Hence, we only perform one iteration to reduce the computational burden and leave the remaining detail enhancement work to the depth-guided deformation stage. We denote the new mesh after updating as $M'_c$. 

A direct way to learn the SDF function from $S_f$ is by applying PIFu-N on $S_f$. However, It may lead to a misalignment between the generated SDF field and the coarse mesh $M_c$, thus challenging the deformation. Therefore, as illustrated in Fig.~\ref{fig:pipeline_fine}, we render $M_c$ into a depth map $D_c$, and feed $D_c$ and $S_f$ together into a Pix2Pix module to infer a normal map $N$ for conducting PIFu-N.

\subsubsection{Depth-guided Mesh Refinement} Although normal-assisted PIFu can model details better than other existing methods, generating details as reflected in the normal map is still practically challenging. Our experiments found that the learned depth maps contain richer geometric details than the learned implicit fields. Thus we propose a depth-guided deformation method to enhance $M'_c$ further. Specifically, as illustrated in Fig.~\ref{fig:pipeline_fine}, we first render $M'_c$ into a depth map $D'_c$ and feed it together with $N$ into a new Pix2Pix module for generating a depth map $D_f$ with sharper details than $D'_c$. Here, we use $N$ instead of $S_f$ since $N$ has already captured the geometric information from $S_f$ and can ease the learning procedure. 

\noindent{\textit{\textbf{Without Depth Alignment.}}} To transfer geometric details from $D_f$ to $M'_c$, a straightforward way is to first convert $D_f$ to a point cloud $P$ and then fit $M'_c$ to $P$. Specifically, for each vertex $\mathbf{v}$ of $M'_c$, we retrieve $K$ closest points in $P$ and employ the inverse distance weighting algorithm~\cite{bartier1996multivariate} to directly update the position of $\mathbf{v}$. 

\noindent{\textit{\textbf{Flow-based Local Depth Alignment.}}} 
Although the design of the method discussed above well guarantees global alignment between $P$ and $M'_c$, there is no guarantee for local alignment. Implicit-guided mesh updating is hard to ensure the alignment of local geometry (e.g., nose) between the $M'_c$ and $S_f$ (thus, both $N$ and $D_f$ may also suffer from misalignment). Directly fitting $M'_c$ to $D_f$ tends to cause artifacts due to the local misalignment between them, as shown in Fig.~\ref{fig:warping}. Multiple iterations and extensive smoothings are required to obtain stable results, which is inefficient and may result in blurry geometric details. 
To address this issue, we propose a flow-based alignment method. More specifically, we train a FlowNet~\cite{ilg2017flownet} to take $D_f$ and $D'_c$ as input and output a warping field. The warping field is applied to align $D_f$ to $M'_c$ and generate an aligned/warped depth $D'_f$. Then a high-quality point cloud $P$ can be extracted from $D'_f$. Thus, $P$ is also locally aligned with $M'_c$. The good alignment between $P$ and $M'_c$ facilitates the registration of local geometric details from $P$ to $M'_c$. As a result, the final mesh $M_f$ is close to $M'_c$ but with more local details, instead of being completely aligned with $S_f$. The alignment of the sketch, depth maps, and normal map used in Fig.~\ref{fig:pipeline_fine} is shown in Fig.~\ref{fig:fine_alignment}. Although a minor global misalignment exists between $M_f$ and $S_f$, the resulting mesh is still plausible and convincing, as illustrated in Fig.~\ref{fig:gallery}. Thanks to the local alignment, we found that one iteration of the depth-guided mesh refinement is enough to reconstruct vivid details stably (the improvement of multiple iterations is slight), reducing the computational cost.

\input{figure/fig_warping}

\input{figure/fig_fine_alignment}

\subsubsection{Training} IDGMM is backed by four learning-based models: Pix2Pix-1 that maps $S_f \oplus D_c$ ($\oplus$ indicates concatenation) to $N$, Pix2Pix-2 that maps $D'_c \oplus N$ to $D_f$, PIFu-N and FlowNet. All the networks are trained separately and in a fully-supervised manner: 1) To train Pix2Pix-1, for each ground-truth mesh $M$ (which contains rich details), we render its ridge and valley lines as input fine strokes, using the tool provided by Suggestive Contours~\cite{decarlo2003suggestive}. The stroke types are encoded by the channel of red or green colors, and the depth is encoded with the shades of the color. Specifically, the ridge is encoded in ($c$, 0, 0) and the valley in (0, $c$, 0), $c = 255 - |k|$, where $k$ is the curvature of a line segment. Thus the smaller value of $c$, the visually greater color depth (i.e., visually darker), representing the higher ridge or deeper valley. In our experiments, the trained model can generalize well to strokes of varying widths, though the strokes in the training set are in a constant width. 2) We smooth $M$ to be $M_s$ and use it as $M_c$ to render depth maps as $D_c$ for training Pix2Pix-1 ($N$ is rendered from $M$). 3) We put $M$ into a $128^3$ SDF field (noted as $g^{128}_M$) and extract the mesh $M_l$. Then we render $M_l$ into a depth map to approximate $D'_c$ for training Pix2Pix-2. 4) We subdivide $M$ to get $M'$ with dense points and deform $M'$ under the guidance of $g^{128}_M$ to generate a new mesh $M_g$. We render $M'$ and $M_g$ to depth maps to approximate $D_f$ and $D'_c$. As $M_g$ and $M'$ are topologically consistent, it is easy to obtain a dense flow as supervision to train FlowNet.

%% file: figure/fig_pipeline_coarse.tex
\begin{figure}[htbp]
  \centering
  \includegraphics[width=.98\linewidth]{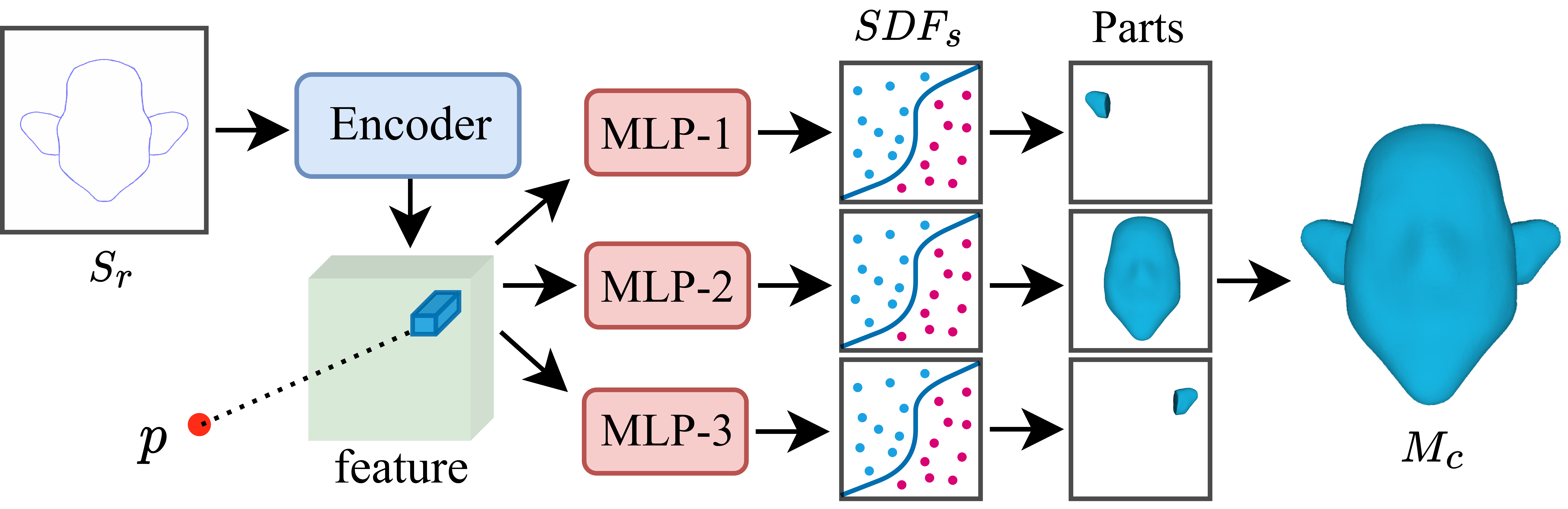}
  \caption{An illustration of o our part-separated coarse modeling of a 3D face with an outline sketch input $S_r$. It shows the generation of three parts of a face region and two ears using PIFu, and then assembles and merges them to obtain a coarse model $M_c$.}
  \label{fig:pipeline_coarse}
\end{figure}


%% file: figure/fig_pipeline_fine.tex
\begin{figure*}[!t]
  \centering
  \includegraphics[width=.93\linewidth]{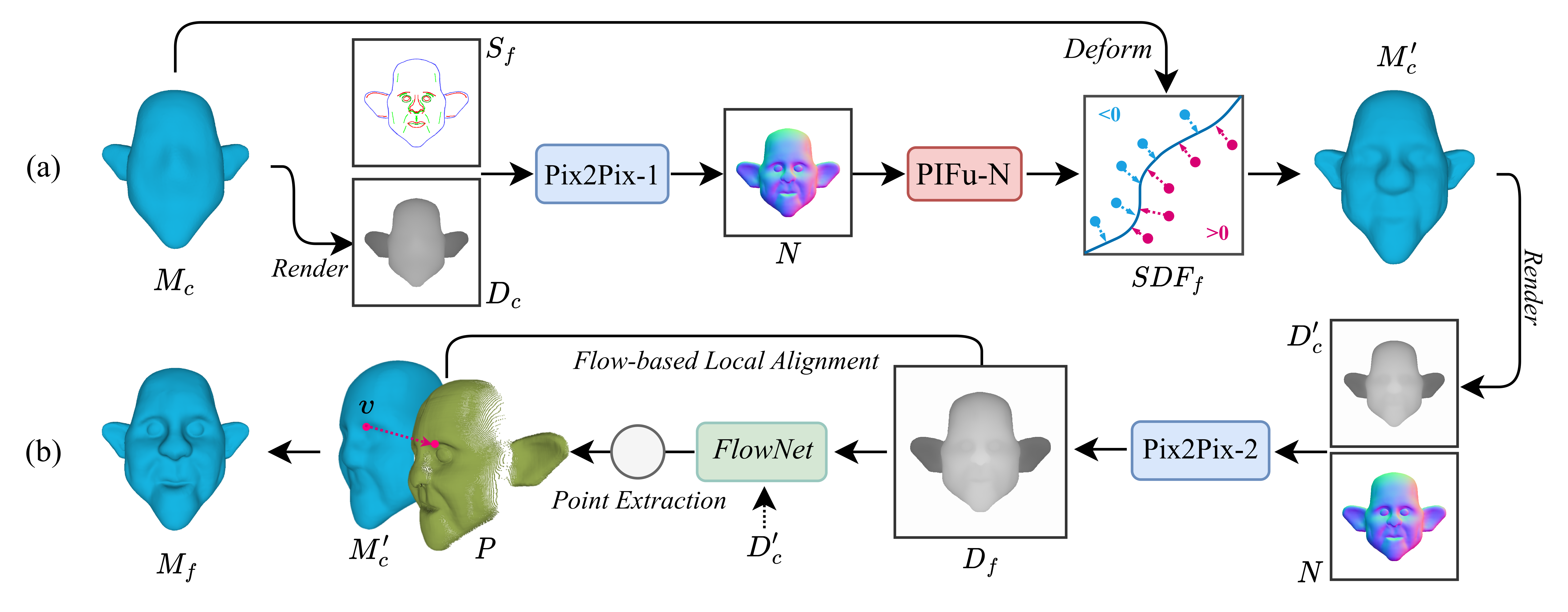}
  \caption{The architecture of our IDGMM. (a) Taking a coarse mesh $M_c$ as input, it is first rendered into a depth map $D_c$. $D_c$ together with the input fine sketch $S_f$ are fed into Pix2Pix-1 to generate a normal map $N$. $N$ is applied to generate an implicit field using PIFu-N. Under the guidance of the SDF field, $M_c$ is deformed to obtain an updated mesh $M'_c$. (b) We then render $M'_c$ into a depth map $D'_c$, which is enhanced to $D_f$ with a Pix2Pix-2 module. After a flow-based local depth alignment, we obtain a high-quality point cloud $P$ from the warped depth map. $P$ is locally aligned with $M'_c$ and used to guide mesh refinement from $M'_c$ to the resulting mesh $M_f$. Note that the process of sketching is iterative, and the mesh obtained at step (n-1) is used as the input $M_c$ for the n-th step.
  }
  \label{fig:pipeline_fine}
\end{figure*}

%% file: figure/fig_warping.tex
\begin{figure}[htbp]
  \centering
  \includegraphics[width=.98\linewidth]{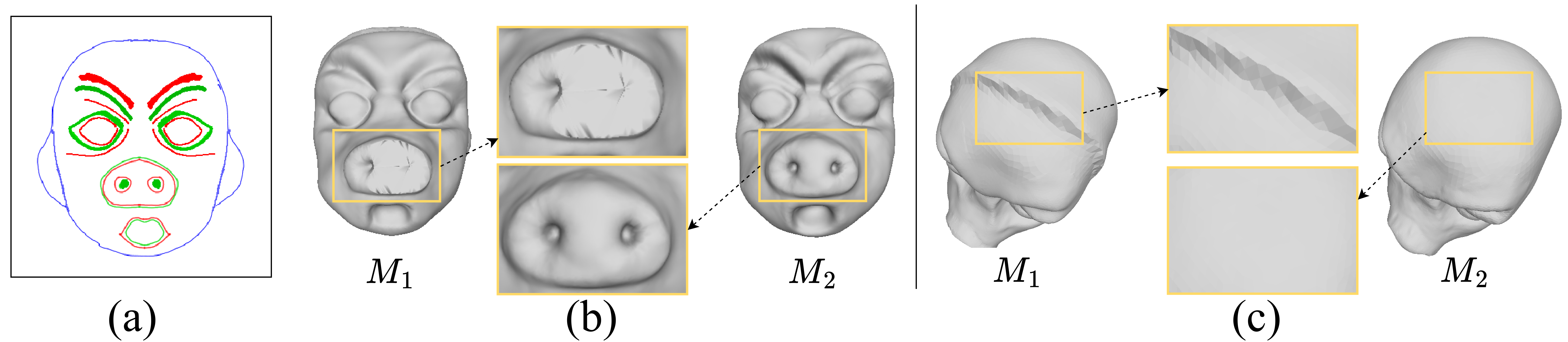}
  \caption{An illustration of results without and with Flow-based Local Depth Alignment. (a) the input sketch. (b) the front view of the results. (c) the top view of the results. Our flow-based alignment ($M_2$) resolves the artifacts caused by directly fitting $M'_c$ to $D_f$ without depth alignment ($M_1$).
  }
  \label{fig:warping}
\end{figure}

%% file: figure/fig_fine_alignment.tex
\begin{figure*}[h]
  \centering
  \includegraphics[width=.98\linewidth]{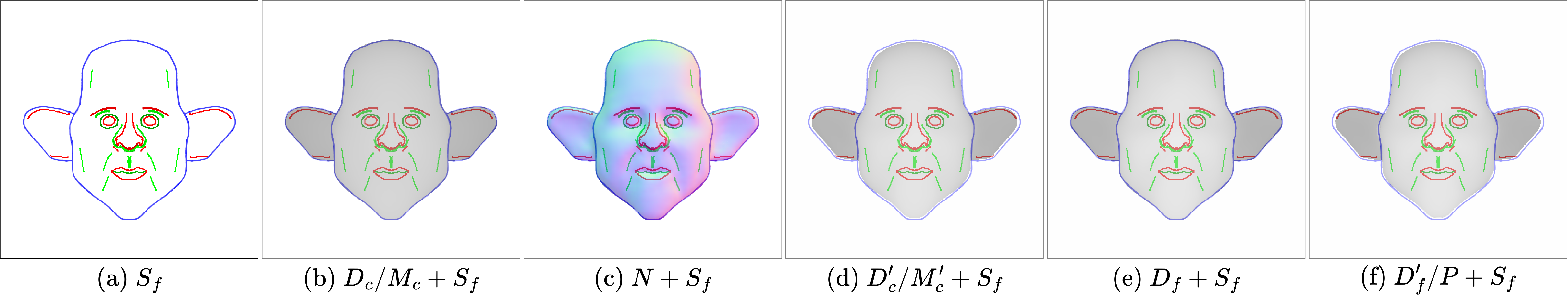}
  \caption{An illustration of the alignment of the sketch, depth-maps, and normal-map used in Fig.~\ref{fig:pipeline_fine}. The overlapping images of $S_f$ and $D_c$, $N$, $D'_c$, $D_f$, $D'_f$ are shown above. Note that $D_c$ is rendered by $M_c$, while $D'_c$ is rendered by $M'_c$. $P$ is extracted from the warped depth (denoted as $D'_f$ here) generated by FlowNet. The resulting mesh $M_f$ of IDGMM is close to $M'_c$ but with more local details, instead of being completely aligned with $S_f$.}
  \label{fig:fine_alignment}
\end{figure*}


%% file: section/results.tex
\section{Results and Evaluation}
\label{sec:results}
In this section, we will evaluate our system from two aspects, namely, \textbf{system usability} (Section ~\ref{result:system}) and \textbf{algorithm effectiveness} (Section ~\ref{result:algorithm}).

\subsection{Evaluation on System Usability}
\label{result:system}

\noindent{\textit{\textbf{Apparatus.}}} Our user interface was implemented with QT and deployed on a desktop workstation with one Intel i5 @2.7GHz CPU and 8GB of memory. Users can interact with the system with a computer mouse or a pen tablet. The neural-network-based backend model was implemented with Pytorch 1.8.1 and deployed on a server with one Intel i7 @4.2GHz CPU, 16 GB of memory, and one NVIDIA GTX 3090 GPU graphics card. To support the training of our proposed algorithms for modeling high-fidelity 3D heads, we merged the existing datasets of 3DAnimalHead~\cite{luo2021simpmodeling} and 3DCaricShop~\cite{qiu20213dcaricshop}, resulting in 4,528 high-quality models in total. Then we split these data into 9:1 for training and testing in our experiments. Please refer to our supplemental materials for the implementation details of the neural networks.

\noindent{\textit{\textbf{Participants.}}} Our key objective is to create a 3D modeling system that is easy to use for amateur users without 3D modeling experience. To verify this, we invited 16 subjects (P1-P16, aged 18 to 32) to participate in this evaluation session, none of whom had experience in 3D modeling. Six of them (P2, P3, P6, P7, P8, P12) had professional 2D drawing experience, and the remaining had limited drawing experience. Before the modeling session, each participant was given 10 minutes to watch an video showing the basic operations of our system. After the tutorial, each user had 20 minutes to get familiar with our system. All the participants were asked to perform comparison and usability studies.

\subsubsection{Comparison Study} We first conducted a comparison study on different modeling systems to demonstrate the superiority of our system. After thoroughly reviewing existing sketch-based character modeling systems, we chose DeepSketch2Face~\cite{han2017deepsketch2face} and SimpModeling~\cite{luo2021simpmodeling} for comparison since these systems can be easily accessed. For DeepSketch2Face, its released system was used. We asked the authors of SimpModeling to provide their system to us. ZBrush is a powerful commercial software for assisting professional artists in creating arbitrary 3D models. We also added ZBrush to our informal comparison on face modeling. For a fair comparison, all 16 subjects were also given 10 minutes to learn through a tutorial and 20 minutes to get familiar with each of the other systems before the formal user study. In the formal session, each user was given a shading image of a 3D model as a reference. She/he was requested to create 3D models referring to the given image using the four compared systems (i.e., DeepSketch2Face, SimpModeling, \textit{SketchMetaFace}, and ZBrush) in random order. Note that all the tools provided by SimpModeling and ZBrush are 3D interactive operations, while most operations of DeepSketch2Face and \textit{SketchMetaFace} focus on the 2D canvas.

Fig.~\ref{fig:system_compare} shows the reference images, the created models with the four systems, and the corresponding modeling time. Compared to DeepSketch2Face and SimpModeling, our system supported users to create more appealing shapes and craft more vivid surface details. The geometric shape and surface details created by our system are closer to the reference models. Compared to ZBrush, our system took less time for users to create visually reasonable 3D models. To complete each model, each user took around 2-5 minutes to use DeepSketch2Face, around 7-15 minutes with SimpModeling, around 5-9 minutes with our system, and around 10-18 minutes with ZBrush. Most participants complained that DeepSketch2Face was hard to use as it could only output human faces (mainly because of the limited parametric space of the human face). They mentioned that SimpModeling could create coarse shapes and some minor details, but it was challenging to learn and use. We observed that most subjects got stuck in the coarse shape modeling process with SimpModeling and ZBrush. Some even gave up adjusting coarse shapes and directly turned to sculpting surface details. ``The 3D operations are difficult to use, and I need to speed time adjusting the shape. I am disappointed with SimpModleing and ZBrush'', as commented by P8. ``3D interactions are extremely unfriendly to me. I need to switch perspectives frequently. These frequent switching operations make me irritable'' (P11). Most subjects enjoyed the modeling process defined by \textit{SketchMetaFace}.
Some participants reported that \textit{SketchMetaFace} was user-friendly and allowed for creating vivid avatar heads easily. They also pointed out that our system saved much time and labor in generating 3D heads. ``\textit{SketchMetaFace} is much better than SimModeling. The coarse shape modeling provided by \textit{SketchMetaFace} is easier and can save me a lot of time. The curvature-aware strokes allow me to craft details freely in an intuitive way'' (P6). ``It is very cool to create 3D models by drawing sketches. I am looking forward to using \textit{SketchMetaFace} in the future.'' P1 suggested that the 3D sculpting tools (e.g., smooth and crease) provided by ZBrush could be added to the fine stage, supporting users in further fine-tuning geometric details.

\input{figure/fig_system_compare}

\subsubsection{Usability Study} In this study, each participant was asked to freely create at least one model without restrictions on result diversity, result quality, or time duration. Fig.~\ref{fig:gallery} shows a gallery of models created by these participants, which reflect the expressiveness of our system. It can be seen from this figure that our system supports amateurs in geometrical modeling to create character faces with diversified shapes and rich geometric details. All of the participants felt that our system was powerful in creating diversified avatar heads, and they were deeply impressed by the simplicity, intuitiveness, and controllability of our system. It is worth mentioning that two of the participants said they enjoyed the process very much and expressed their desire to learn 3D modeling. 

\input{figure/fig_gallery}

Most of the participants liked the intuitive stroke suggestion tool, which was quite helpful for them in figuring out the meaning of curvature-aware strokes. We observed that the participants with great drawing skills (i.e., P2, P3, P6, P7, P8, and P12) quickly became used to working with the curvature-aware strokes thanks to the suggestion tool. Once grasping curvature-aware strokes, they preferred to paint each part of the model from scratch and customize desired details by themselves instead of searching for a specific structure using the stroke suggestion module. P6 commented ``The stroke suggestion tool is a very nice and useful function for assisting me in understanding the usage of curvature-aware strokes.'' We received similar positive comments from P7 and P12: ``With the help of the stroke suggestion function, I can easily understand how to depict geometric structures using curvature-aware strokes'' (P7); ``The curvature-aware strokes are useful and powerful for carving models' details, like wrinkles'' (P12). Other participants tended to use the stroke suggestion function throughout the whole modeling process due to their limited drawing skills. ``The suggestion module is easy and intuitive to use. I do not need to spend much time thinking about how to paint a correct sketch. It avoids frequent modifying operations'' (P1). ``The suggestion module is convenient and friendly for me. It reduces a lot of manual operations and allows me to create diversified results in a very easy way'' (P5). ``I can make funny and realistic results by simply searching and integrating different parts in minutes (two eyes, a nose, and a mouth)'' (P10). 

The participants also provided some constructive comments. For example, P4 said, ``It would be better to allow me to search for a suitable head contour in the coarse modeling stage, just like searching for a nose or a mouth in the fine stage.'' One potential solution is collecting a coarse shape database and applying the retrieval mechanism in the coarse-shape modeling stage. ``Although the profile depth editing tool allows me to adjust models in the side view, the system still fails to create an elephant's nose. I do not know how to create an elephant's nose using the tools provided by \textit{SketchMetaFace}.'' said P2. Enlarging our datasets and adopting multi-view drawing in the coarse stage would be a possible solution for this problem.

\subsubsection{Questionnaire Analysis}  At the end of the comparison study, each participant was required to complete a System Usability Scale (SUS) questionnaire and a NASA Task Load Index (NASA-TLX) questionnaire to evaluate the usability and workload of our system. We found that the overall SUS score of our system was 79, out of a scale of 100 (DeepSketch2Face: 64, SimpModeling: 38, ZBrush: 41), indicating the good usability of our system~\cite{bangor2009determining}. In Fig.~\ref{fig:chart}(a), we show the mean scores for all the individual SUS questions. For the questions with the odd numbers, the higher the SUS scores, the better; for the rest of the questions, the lower the SUS scores, the better. The scores of Q1 and Q9 suggest that the participants appreciated our system and were satisfied with the models created by our system. From Q2-4, Q7-8, and Q10, we can conclude that our system supported amateur users creating desired 3D head models easily and intuitively, indicating the good user efficiency and usability of our system. The scores of Q5-6 show that the participants also recognized our system's well-designed modeling pipeline and tools. Although the high scores of Q3 and Q7 indicate that DeepSketch2Face is easy to use, the participants were disappointed with its results, leading to low scores for Q1 and Q9. The high scores of Q2, Q4, Q6, Q8, and Q10 and the low scores of Q3, Q7, and Q9 all suggest that SimpModleing and ZBrush are unfriendly for these amateur uses. Grasping these two systems is extremely hard for them.

\input{figure/fig_chart}

Fig.~\ref{fig:chart}(b) illustrates the average score for each question in the NASA-FLX questionnaire. The results of our systems are also positive. Compared to SimpModeling and ZBrush, our system's mental demand, physical demand, temporal demand, effort, and frustration are at an extremely low level. It implies that our system does not require users to pay a lot of concentration and effort when using it. The higher performance score of our system reflects that the participants were also more satisfied with their modeling results with our system. The lower performance score and the higher frustration score of SimpModeling and ZBrush than those of our system suggest that it was hard for the participants to create desired results using 3D operations. The lower performance score of DeepSketch2Face demonstrates that the participants were unsatisfied with the results generated by its algorithm, which also leads to a high frustration level.

We conducted a subjective user study to evaluate the faithfulness (i.e., the degree of fitness to reference images/models) of synthesized results. We randomly chose a set of results from the comparison study, containing 15 reference models and the corresponding results created by the participants using the four above systems. We invited 50 subjects to participate in this subjective evaluation through an online questionnaire. Most subjects had no 3D modeling experience, and none had participated in the previous studies. We showed the participants five images for each case (15 cases in total), including the input sketch and the four modeling results by the compared systems, placed side by side in random order. Each participant was asked to score each result based on the faithfulness to the reference model (1 denoting the lowest fitness and 10 for the highest fitness). Fig.~\ref{fig:chart}(c) shows the mean score of each system for this study. This figure shows that the 3D models created by amateurs with our system in the comparison study received relatively higher marks than the counterpart systems, implying that our system could assist novice users in creating desired 3D heads. Statistical analysis also showed that the scores significantly differed across the compared systems. Specifically, we ran Shapiro-Wilk normality tests on the collected data and found non-normality distributions ($p<0.001$). We thus conducted Kruskal-Wallis tests on the faithfulness scores and found significant effects. Paired tests between our system and each of the compared ones confirmed that our system (mean: 6.28) could effectively support amateurs in creating significantly more faithful results to the reference models than the other systems, i.e., DeepSketch2Face (mean: 1.96, $p < 0.001$), SimpModeling (mean: 3.64, $p < 0.001$) and ZBrush (mean: 5.82, $p = 0.008$). More details can be found in our supplementary material.

\subsection{Evaluation on Algorithm Effectiveness}
\label{result:algorithm}

\noindent{\textit{\textbf{Comparison on Part-separated Mesh Inference.}}} There are some alternative methods~\cite{groueix2018papier,wang2015sketch,wang2018pixel2mesh} for inferring part-separated meshes from an input sketch. To verify the generalization ability of part-separated PIFu, we choose two representative alternative methods for comparison. One is a retrieval-based method~\cite{wang2015sketch}, denoted as Retrieval and the other one is a deformation-based method~\cite{wang2018pixel2mesh}, denoted as Pixel2Mesh. The qualitative comparisons are presented in Fig.~\ref{fig:base_level}, where we can see that our results align much better with the input sketches.

\input{figure/fig_base_level.tex}

\noindent{\textit{\textbf{Comparisons on Sketch2Mesh.}}} The core problem of our system is to learn the mapping from $S_f$ to a detailed mesh. To evaluate the superiority of IDGMM, we selected four existing representative methods for comparison: 3D-R2N2~\cite{choy20163d}, Pixel2Mesh~\cite{wang2018pixel2mesh}, DeepSDF~\cite{park2019deepsdf} and PIFuHD~\cite{saito2020pifuhd} (the method used by SimpModeling). All these methods took $S_f$ and $D_c$ as input for fairness. Fig.~\ref{fig:algo_compare} and Tab.~\ref{tab:quant_fine} show the results of this comparison. Both qualitative and quantitative results demonstrate the superiority of our method. Although PIFuHD performs not badly on quantitative measurements, the qualitative results show that our proposed algorithm (IDGMM) performs much better than PIFuHD on geometric details synthesis. Meanwhile, PIFuHD requires a time-consuming mesh extraction process from an implicit field (around 5.0s for one result generation). SimpModeling slightly reduces PIFuHD's time consumption by sampling points along the normal directions and applying local Laplacian deformation (1.0s for one result generation). Our IDGMM combines the advantages of mesh, continuous SDF, and depth map representations, making it very powerful not only in generating detailed 3D geometry but also in inference efficiency (around 0.5s for one result generation).

\input{figure/fig_algo_compare.tex}

\input{table/tab_fine_level_comparison}

\input{figure/fig_ablation_idgmm}

\input{figure/fig_stroke_ablation.tex}

\noindent{\textit{\textbf{Ablation Study on Implicit/Depth Guidance.}}} There are two key components in our proposed IDGMM: implicit-guided mesh updating and depth-guided mesh refinement. To verify the indispensability of these two modules, we compared IDGMM with two alternative settings: 1) without implicit guidance - we use $D_c$ and $N$ as input to generate $D_f$ and corresponding warped $P$, which is then used to guide the deformation from $M_c$. 2) without depth guidance, i.e., $M'_c$ shown in Fig.~\ref{fig:pipeline_fine}. Qualitative results are shown in Fig.~\ref{fig:ablation_idgmm}. The resulting meshes with both implicit and depth guidance outperform the other two options on surface detail generation, implying the necessity of the implicit-guided and depth-guided modules.

\noindent{\textit{\textbf{Ablation Study on Curvature-aware Strokes.}}} The common option to represent sketches is using strokes without any curvature-aware attributes (e.g., DeepSketch2Face and SimpModeling), which is hard to depict complex surface details, as seen in the left part of Fig.~\ref{fig:black_stroke}. The right part of Fig.~\ref{fig:black_stroke} shows the great capability of curvature-aware strokes in representing rich geometric details.

\noindent{\textit{\textbf{Perceptive Evaluation Study.}}} To further evaluate the effectiveness and superiority of our proposed algorithm (part-separated PIFu and IDGMM), we conducted another perceptive evaluation study. We selected 10 samples from the experiments of Comparison on Part-separated Mesh Inference (like Fig.~\ref{fig:base_level}), Comparisons on Sketch2Mesh (like Fig.~\ref{fig:algo_compare}), and Ablation Study on Implicit/Depth Guidance (like Fig.~\ref{fig:ablation_idgmm}) respectively, resulting in three questionnaires. Each case in the questionnaires showed the input sketch and the results generated by different algorithms, placed side by side in random order. The 50 subjects mentioned above were also asked to evaluate each synthesized model's faithfulness (i.e., the degree of fitness to input sketches) on a ten-point Likert scale (1 = lowest fitness to 10 = highest fitness). Fig.~\ref{fig:chart}(d) shows that the results generated by part-separated PIFu fit the input sketches better than Retrieval and Pixel2Mesh. Fig.~\ref{fig:chart}(e) suggests that IDGMM could synthesize richer, more vivid, and more realistic geometric details than the other methods. Fig.~\ref{fig:chart}(f) indicates the necessity and superiority of combining implicit and depth guidance for detailed geometry generation. For statistical analysis, we first performed Shapiro-Wilk normality tests, respectively, for the three collected data and found that all of them followed non-normality distributions ($p < 0.001$). Therefore, we conducted a Kruskal-Wallis test on the faithfulness scores for each perceptive evaluation, and the results also showed significance across different comparisons. 
For the evaluation of coarse shape modeling, paired tests showed that our method (mean: 8.60) performs significantly better on diverse shape generation than both Retrieval (mean: 3.85, $p < 0.001$) and Pixel2Mesh (mean: 5.38, $p < 0.001$). 
For the evaluation of surface detail generation, the results indicated that IDGMM (mean: 8.90) led to significantly more faithful results than the other methods, i.e., 3D-R2N2 (mean: 3.25, $p < 0.001$), Pixel2Mesh (mean: 3.89, $p < 0.001$), DeepSDF (mean: 5.43, $p < 0.001$), and PIFuHD (mean: 6.63, $p < 0.001$).
For the evaluation of implicit/depth guidance, the tests suggested that depth\&implicit guidance (mean: 8.55) significantly performs better on geometric detail synthesis than the alternative options, i.e., only implicit guidance (mean: 6.23, $p < 0.001$) and only depth guidance  (mean: 5.95, $p < 0.001$). It is worth mentioning that the difference between depth and implicit guidance was not distinct ($p = 0.169$). This is consistent with our expectation, since both only using depth refinement and only using implicit refinement can synthesize minor details. But they fail to depict high-quality geometric details, further confirming the significant positive effect of incorporating implicit and depth refinement. All these statistical results confirmed that all our proposed algorithms significantly outperform the corresponding alternative options. More details about evaluation are provided in our supplementary material.


%% file: figure/fig_system_compare.tex
\begin{figure*}[!t]
  \centering
  \includegraphics[width=.96\textwidth]{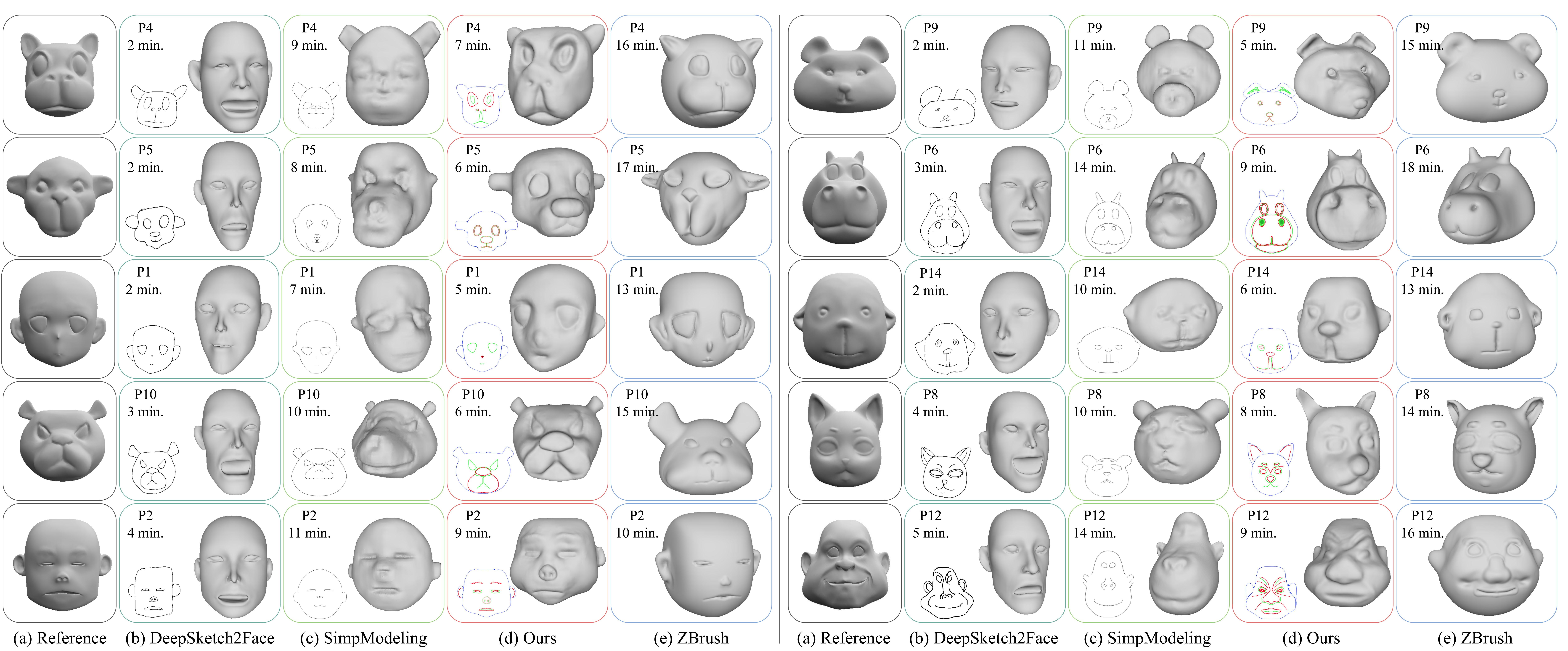}
  \caption{Comparison between our system against the state of the arts. The results in each row were created by the same user given a reference in (a). For each system, we show the sketch, resulting model, drawing time, and the corresponding participant.
  }
  \label{fig:system_compare}
\end{figure*}


%% file: figure/fig_gallery.tex
\begin{figure*}[htbp]
  \centering
  \includegraphics[width=0.92\linewidth]{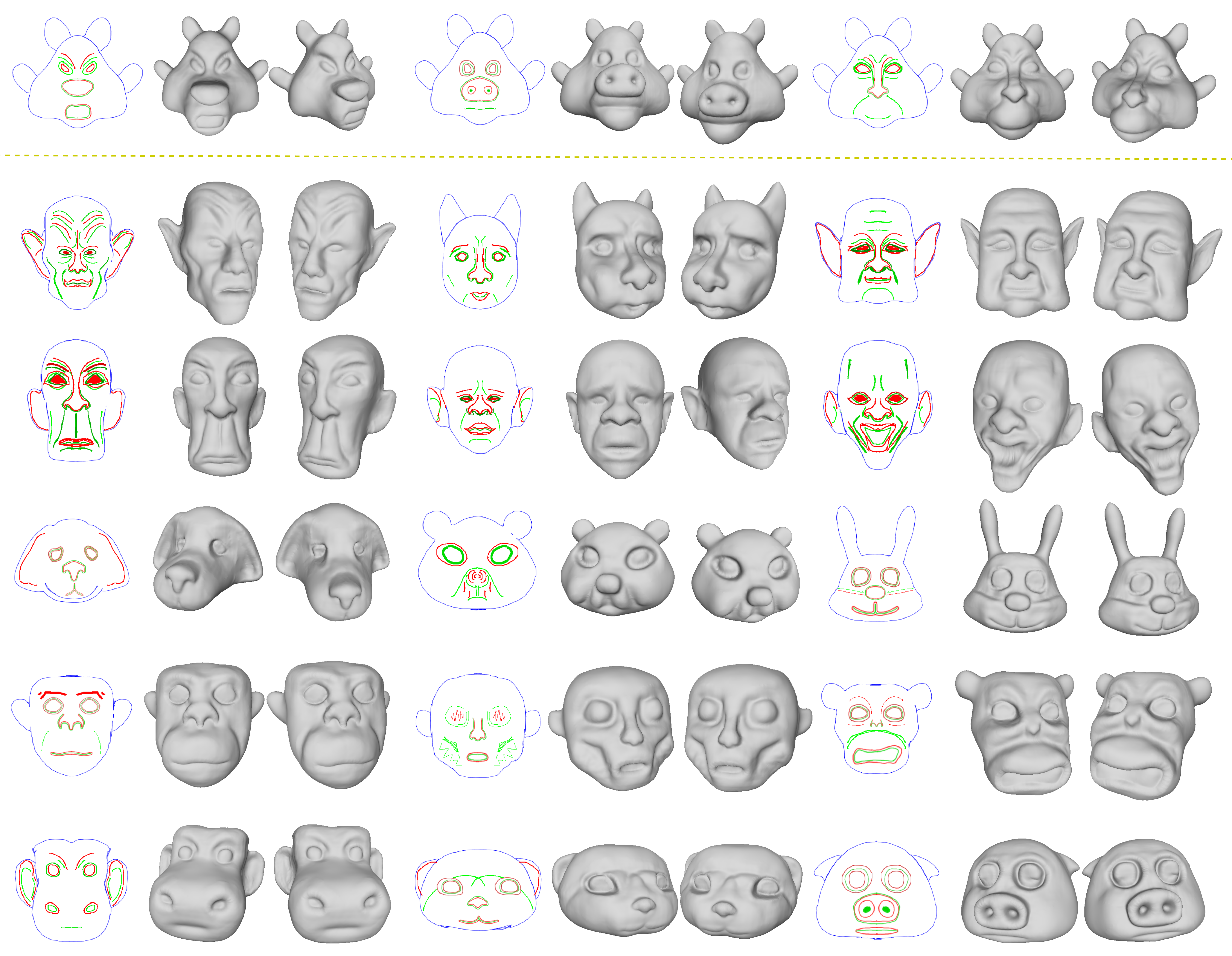}
  \caption{The gallery of our results. All models are created by amateur users who are trained to use our system with a tutorial. Thanks to the easy-to-use two-stage modeling design and the stroke suggestion component, the users can complete each model design in 5-9 minutes. The three results in the first row were created using the same coarse mesh but applying different surface details.}
  \label{fig:gallery}
\end{figure*}

%% file: figure/fig_chart.tex
\begin{figure*}[htbp]
  \centering
  \includegraphics[width=0.98 \textwidth]{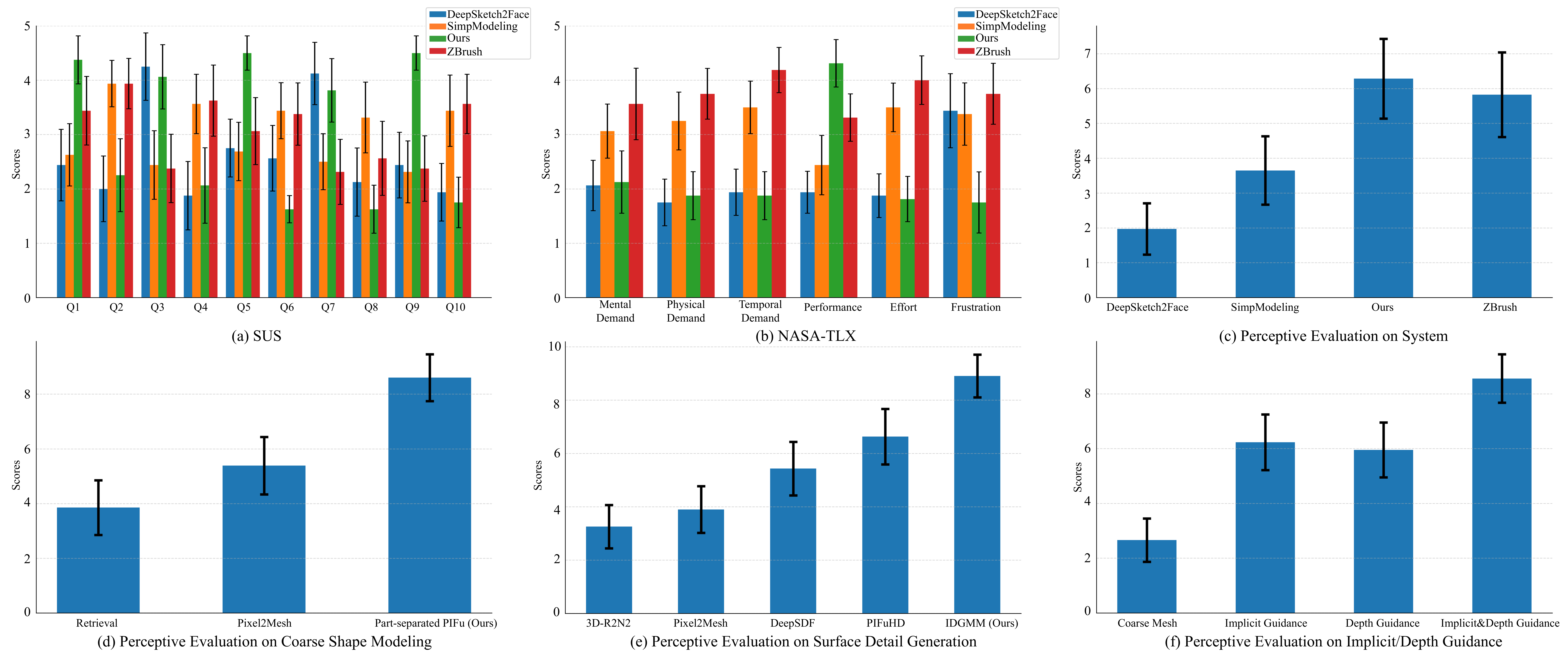}
  \caption{(a) Mean scores of SUS in a 5-point scale. (b) Mean scores of NASA-TLX in a 5-point scale. (c) Perceptive evaluation on results of the compared systems. (d) Perceptive evaluation on coarse shape modeling. (e) Perceptive evaluation on surface detail generation. (f) Perceptive evaluation on implicit/depth guidance. Each error bar represents the standard deviation of the corresponding mean.}
  \label{fig:chart}
\end{figure*}

%% file: figure/fig_base_level.tex
\begin{figure}[!t]
  \centering
  \includegraphics[width=0.85\linewidth]{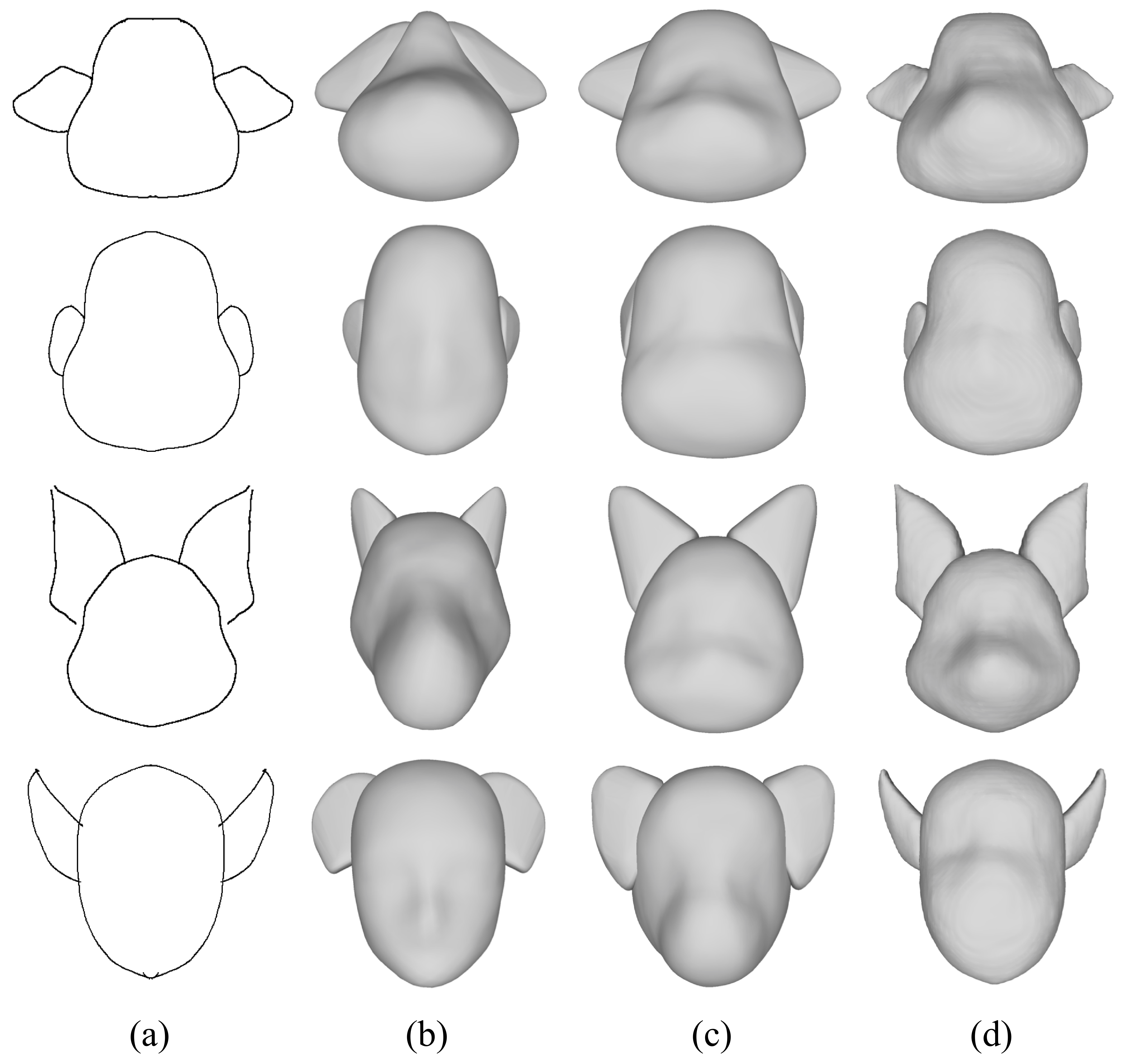}
  \caption{Qualitative comparisons on part-separated mesh inference from an input sketch (a). (b) The results of retrieval. (c) The results of Pixel2Mesh. (d) The results of our part-separated PIFu.
  }
  \label{fig:base_level}
\end{figure}


%% file: figure/fig_algo_compare.tex
\begin{figure}[!htbp]
  \centering
  \includegraphics[width=.95 \linewidth]{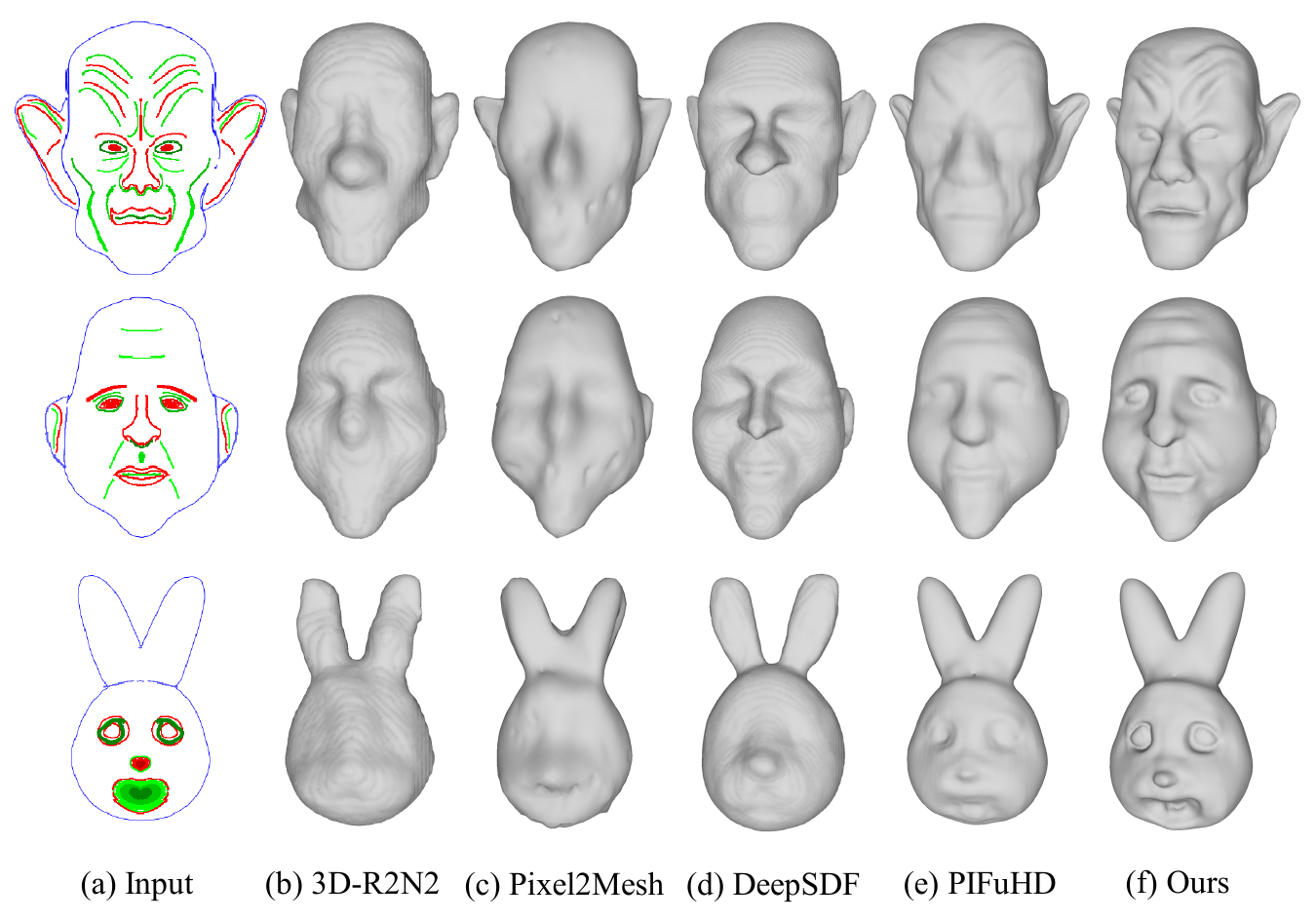}
  \caption{Qualitative comparisons of our IDGMM with four existing methods for Sketch2Mesh inference.}
  \label{fig:algo_compare}
\end{figure}



%% file: table/tab_fine_level_comparison.tex

\begin{table}
    \footnotesize
    \caption{Quantitative comparison with our proposed IDGMM with four existing methods for Sketch2Mesh inference. We adopt IoU, Chamfer-$L_2$, and normal consistency to evaluate the results.}
    \label{tab:quant_fine}
    \begin{center}
        \begin{tabular}{l||c|c|c}
            & IoU $\uparrow$   & Chamfer-$L_2$ ($\times 10^2)$ $\downarrow$  & Normal-Consis. $\uparrow$ \\ \hline \hline 
            3D-R2N2 & 0.858 & 0.149 & 0.929 \\  
            Pixel2Mesh & 0.882 & 0.123 & 0.937 \\ 
            DeepSDF & 0.894 & 0.117 & 0.949 \\ 
            PIFuHD & 0.911 & 0.103 & 0.955 \\
            Ours & \textbf{0.915} & \textbf{0.099} & \textbf{0.956} \\
        \end{tabular}
    \end{center}
\end{table}

%% file: figure/fig_ablation_idgmm.tex
\begin{figure}[!t]
  \centering
  \includegraphics[width=.92\linewidth]{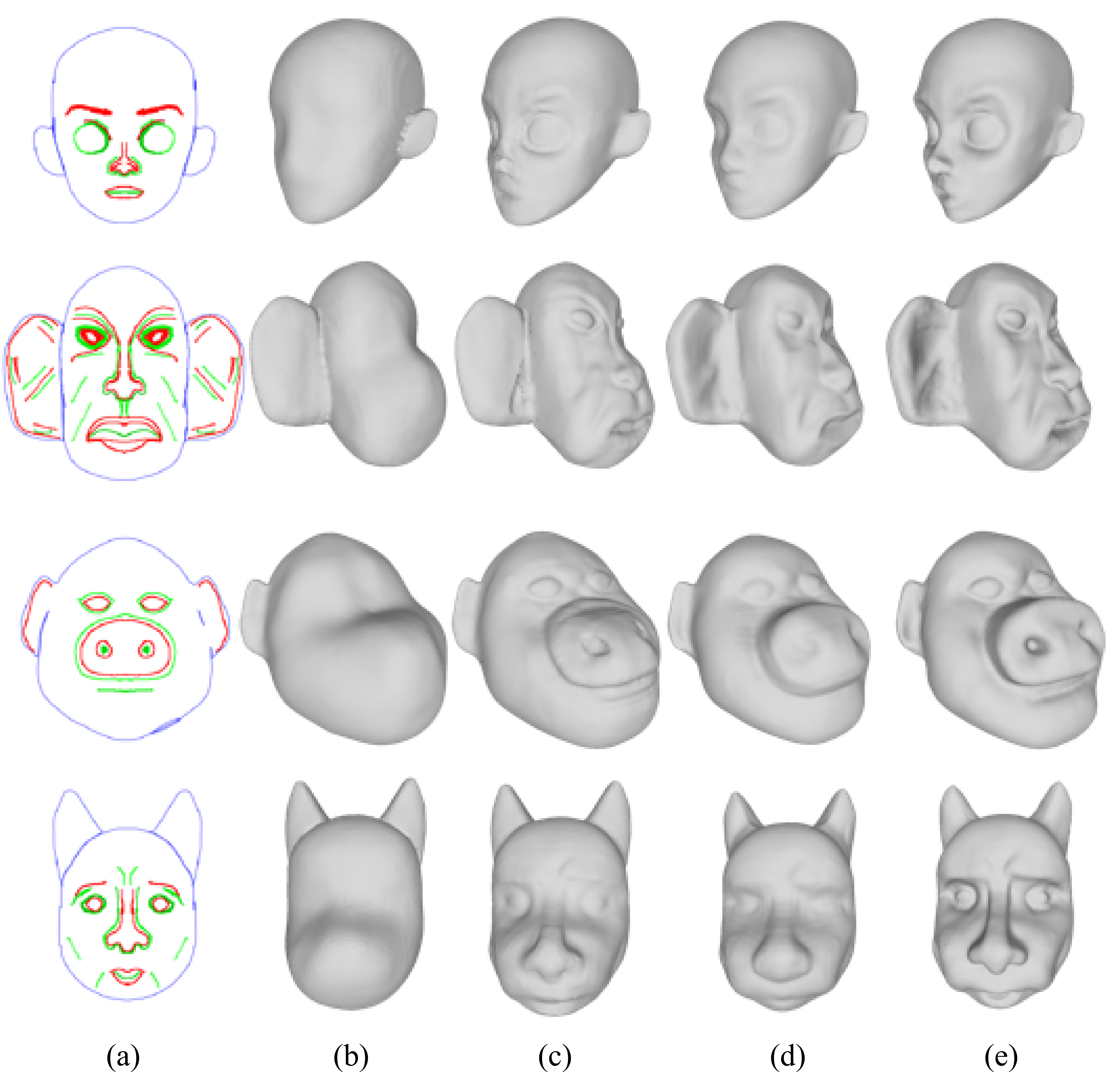}
  \caption{Ablation study on implicit/depth guidance. From left to right: (a) input sketch; (b) coarse mesh (i.e., $M_c$ in Fig.~\ref{fig:pipeline_fine}); (c) resulting mesh with only depth guidance (without implicit guidance); (d) resulting mesh with only implicit guidance (without depth guidance, i.e., $M'_c$ in Fig.~\ref{fig:pipeline_fine}); (e) resulting mesh with both guidance (i.e., $M_f$ in Fig.~\ref{fig:pipeline_fine}).}
  \label{fig:ablation_idgmm}
\end{figure}

%% file: figure/fig_stroke_ablation.tex
\begin{figure}[!t]
  \centering
  \includegraphics[width=.92\linewidth]{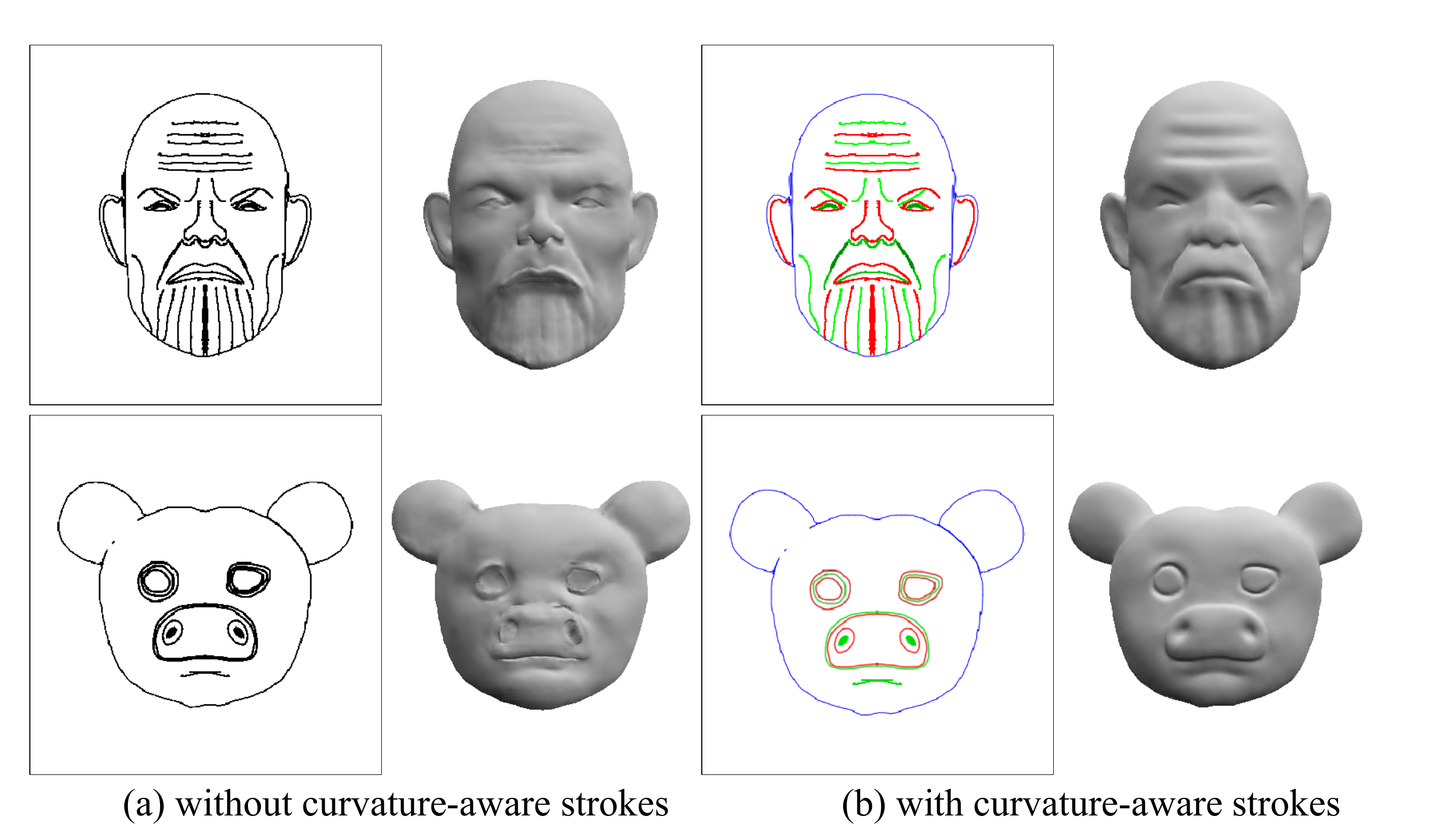}
  \caption{Ablation study on without/with curvature-aware strokes. Using curvature-aware strokes significantly helps enhance the quality of the generated geometric details.}
  \label{fig:black_stroke}
\end{figure}


%% file: section/conclusion.tex
\color{black}
\section{Conclusion}
In this paper, we presented an easy-to-use sketching system for amateur users to create and high-fidelity 3D face models. Both the user interface and the algorithm are carefully designed. Firstly, curvature-aware strokes are utilized to assist users in easily carving geometric details. Secondly, a coarse-to-fine interface is designed. In the coarse stage, users only need to model face contours and the 3D layout of ears. Then, in the fine stage, all interactions are operated on a 2D canvas for detail drawing. Thirdly, to support the accuracy and usability of the user interface, a novel method, named Implicit and Depth guided Mesh Modeling (IDGMM), is proposed. It combines the advantages of the implicit (SDF), mesh, and depth representations, and reaches a good balance between output quality and inference efficiency. Both evaluations of the system and algorithm demonstrate that our system is of better usability than existing systems and the proposed IDGMM also outperforms existing methods.  

\input{figure/fig_limitation.tex}

Although our system is able to create 3D models with diversified shapes and rich details, it also has some limitations (Fig.~\ref{fig:limitation}): a) As we only focus on frontal-view sketching for detail carving, some organs with complex depth changing are hard to model, such as the nose of an elephant; b) When the strokes are densely placed, it cannot produce reasonable geometric details as a large number of vertices are required in this scenario, which our current system does not support. In the future, we will enlarge our dataset to support users in modeling shapes with other categories, such as cartoon character bodies and human garments. We will also try to take multi-view sketches as input to further support the creation of complex models, such as elephants. Meanwhile, we will explore the possibilities to carve high-resolution models efficiently and support richer detail crafting effectively.

\noindent\textbf{Acknowledgements.} 
The work was supported in part by NSFC-62172348, the Basic Research Project No. HZQB-KCZYZ-2021067 of Hetao Shenzhen-HK S\&T Cooperation Zone, the National Key R\&D Program of China with grant No. 2018YFB1800800, the Shenzhen Outstanding Talents Training Fund 202002, the Guangdong Research Projects No. 2017ZT07X152 and No. 2019CX01X104, the Guangdong Provincial Key Laboratory of Future Networks of Intelligence (Grant No. 2022B1212010001), the Shenzhen Key Laboratory of Big Data and Artificial Intelligence (Grant No. ZDSYS201707251409055), and the Key Area R\&D Program of Guangdong Province with grant No. 2018B030338001. It was also partially supported by Outstanding Yound Fund of Guangdong Province with No. 2023B1515020055, Shenzhen General Project with No. JCYJ20220530143604010, Hong Kong Research Grants Council under General Research Funds (HKU17206218), grants from the Research Grants Council of the Hong Kong Special Administrative Region, China (No. CityU 11212119) and the Centre for Applied Computing and Interactive Media (ACIM) of School of Creative Media, CityU.

%% file: figure/fig_limitation.tex
\begin{figure}[!t]
  \centering
  \includegraphics[width=.93\linewidth]{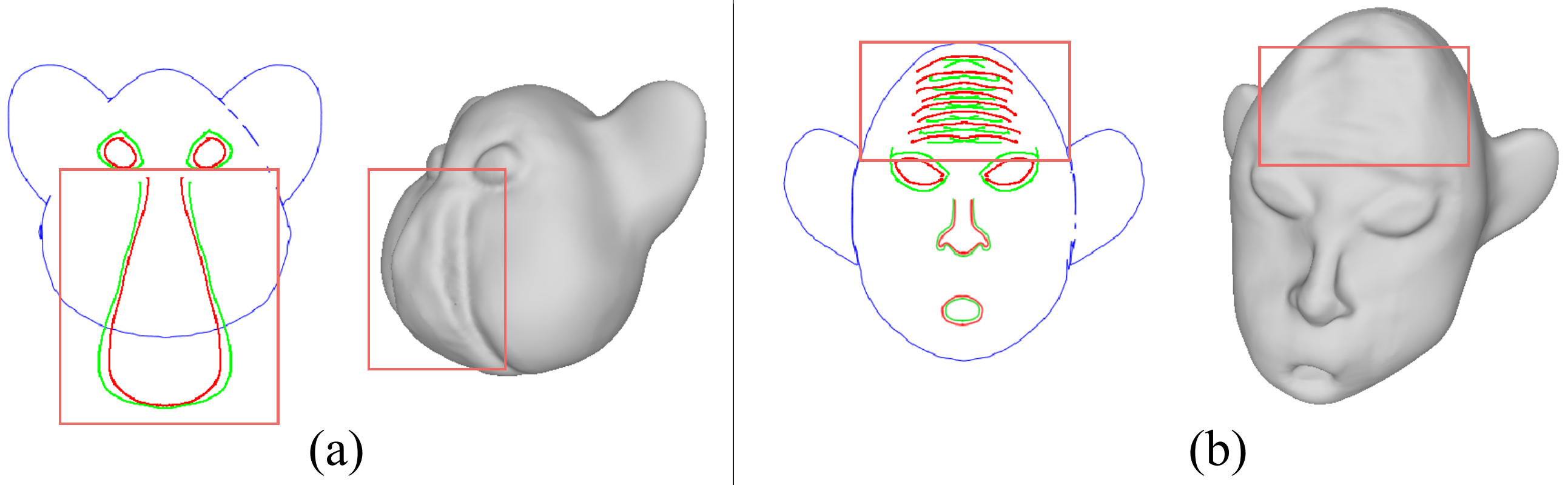}
  \caption{Limitations of our system. Our system also suffers from limitations when a) modeling facial components or details with complex depth changes; b) strokes are placed too densely.
  }
  \label{fig:limitation}
\end{figure}

